\documentclass[conference]{IEEEtran}
\IEEEoverridecommandlockouts
\usepackage{cite}
\usepackage{amsmath,amssymb,amsfonts}
\usepackage{algorithmic}
\usepackage{graphicx}
\usepackage{textcomp}
\usepackage{xcolor}
\def\BibTeX{{\rm B\kern-.05em{\sc i\kern-.025em b}\kern-.08em
    T\kern-.1667em\lower.7ex\hbox{E}\kern-.125emX}}

\usepackage{multirow}
\usepackage{bm}
\usepackage{copyrightbox}
\usepackage{etoolbox}
\newcommand*\rot{\rotatebox{90}}

\def\tablename{Table}

\def\ie{{\it i.e}.} 
\def\eg{{\it e.g}.} 
\def\etal{{\it et al.}}

\def\para#1{\smallskip\noindent{\bf{#1}}}

\newcommand{\specialcell}[2][c]{%
  \begin{tabular}[#1]{@{}c@{}}#2\end{tabular}}

\newcommand{\thisheight}{0.24\columnwidth}
\newcommand{\thiswidth}{\thiswidth}

\begin{document}

\title{\LARGE \bf
3D Pipe Network Reconstruction Based on Structure from Motion with
Incremental Conic Shape Detection and Cylindrical Constraint
}

\author{Sho Kagami$^{1}$, Hajime Taira$^{1}$, Naoyuki Miyashita$^{2}$, Akihiko Torii$^{1}$, and Masatoshi Okutomi$^{1}$
\\[5pt]
\textit{1 Dept. of Systems and Control Engineering, Tokyo Institute of Technology} \\
\textit{2 R\&D Group, Olympus Corporation}
}

\maketitle
\thispagestyle{empty}
\pagestyle{empty}

\begin{abstract}
\label{sec:abst}
Pipe inspection is a critical task for many industries and infrastructure of a city.
The 3D information of a pipe can be used for revealing the deformation of the pipe surface and position of the camera during the inspection.
In this paper, we propose a 3D pipe reconstruction system using sequential images captured by a monocular endoscopic camera.
Our work extends a state-of-the-art incremental Structure-from-Motion (SfM) method to incorporate prior constraints given by the target shape into bundle adjustment (BA).
Using this constraint, we can minimize the scale-drift that is the general problem in SfM.
Moreover, our method can reconstruct a pipe network composed of multiple parts including straight pipes,  elbows, and tees. 
In the experiments, we show that the proposed system enables more accurate and robust pipe mapping from a monocular camera in comparison with existing state-of-the-art methods. 
\end{abstract}

\section{Introduction \label{sec:intro}}
\noindent
The needs for pipe inspection are rapidly increasing in various plants, such as chemical refineries, gas distribution, and sewer maintenance. 
Since the damage and clogging in the pipe are substantially related to disastrous failures of the whole system, periodic industrial inspection is necessary to keep its function.
An industrial endoscope, or an industrial videoscope, is commonly used to inspect the inside of a pipe, since it is impossible for people to directly access the inspection site such as gas network underground or inside the building structures. 
When an operator inserts the probe inside the pipe and then visually inspects on a remote screen to reveal the deformations or defects, they need to guess the defect location and 3D structures around the camera only from the 2D image sequences. 

\begin{figure}[t]
  \centering
  \includegraphics[width=\columnwidth]{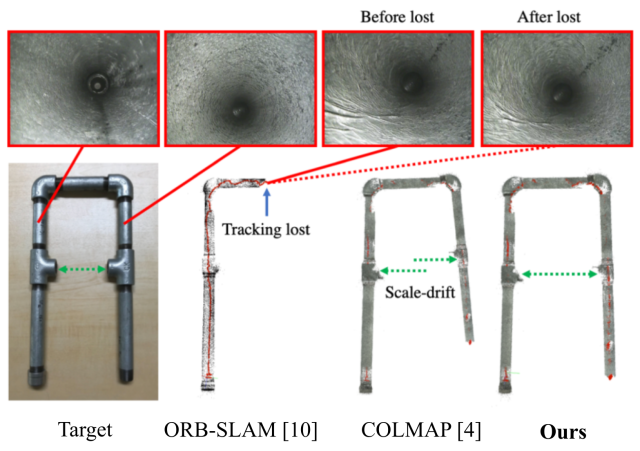}
  \caption{{\bf Difficulties of pipe reconstruction.} Narrow and repetitive structures inside of a pipe make vision-based 3D reconstruction extremely difficult. The reconstruction from such difficult conditions leads to severe drift errors and failure of tracking. We make a SfM system tailored to the pipe structure.\label{fig:difficulty}}
\end{figure}

Vision-based 3D reconstruction techniques such as Structure-from-Motion (SfM)~\cite{photo_tourism,building_roma,visualsfm,colmap,hierarchical,crandall2011discrete,wilson2014robust,sweeney2015optimizing,hsfm} and Visual Simultaneous Localization and Mapping (Visual SLAM)~\cite{orbslam,dso,lsdslam} can potentially help those situation, by reconstruting the 3D structure of the pipe from images, and localizing the probe trajectory during the operation. 
However, because of extraordinarily repetitive and narrow structures as shown in Fig.~\ref{fig:difficulty}, existing methods, which are mostly tested on the urban situation, fail to reconstruct the scene or produce an erroneous structure.

In this paper, we propose a incremental SfM system for a pipe network, with a monocular camera used for the inspection. 
To address the error arisen from the challenging appearances and narrow geometries of the industrial parts, we use prior information of the pipe (assuming a constant inner diameter). 
In contrast to relevant works~\cite{peter2015ijrr,el2011robio}, our system is carefully designed to incorporate the prior constraint into the iterative process of SfM, which enables the system to deal with large errors of the reconstruction and to be capable for the pipe network consisting of multiple straight pipes. 
During the reconstruction, a temporary 3D model is incrementally updated regarding feature correspondences between the previous model and the current image frame from the camera. 
Considering the potential errors induced by scale-drifting, we find multiple straight pipes in the temporary model as conic shapes of the structure. 
Subsequently, our proposed cylindrical constrained BA incrementally refines each pipe to be in line with the given prior. 
Experiments on practical pipe networks consisting of multiple different parts clearly show that our methods obtain more accurate reconstruction compared with the state-of-the-art vision-based methods.

\section{Related work \label{sec:related_work}}
%
\subsection{Vision-based 3D reconstruction \label{sec:vision_recon}}
\noindent
For a set of images input, SfM depicts the captured scene by 3D scene points, based on local feature matching~\cite{sift,dsp_sift} and multi-view triangulation~\cite{mvg,hartley1997triangulation,snavely2008scene}, while estimating the camera poses for the input images. 
In the recent decade, a variety of SfM strategies, including incremental~\cite{photo_tourism, building_roma, visualsfm, colmap}, hierarchical~\cite{hierarchical}, global~\cite{crandall2011discrete, wilson2014robust, sweeney2015optimizing}, and hybrid approach of them~\cite{hsfm} have been studied. 
For a sequential image series input, incremental SfM is the most popular strategy that can be extended to the real-time application~\cite{bronte2014real}. 
On the other hand, SLAM-based methods have been developed in contexts of the real-time operation of estimating the camera trajectory while reconstructing the environment. ORB-SLAM~\cite{orbslam} speeds up the feature extraction process by using binarized ORB descriptor~\cite{orb}. Direct methods~\cite{dso,lsdslam} allow more efficient operation directly obtaining camera trajectory by minimizing a cost regarding the differences of image intensities. 

One common issue of those vision-based reconstruction methods is the accumulated scale-drift problem that causes a camera trajectory and a 3D model inaccurate. 
Several works address the issue via loop closure~\cite{konolige2008frameslam,mei2009constant,strasdat2010vi} that attempts to detect a camera path loop, \eg, using image appearance~\cite{filliat2007icra, galvez2012tor}, and hence detect the scale-drift during the reconstruction. The 3D points and camera poses of the model are then refined as the model keeps consistent scene geometries at the former and the latter of the camera path loop. 
Another approach is to compensate scale information via pre-trained deep architectures that estimate an absolute scale depth~\cite{yang2018eccv, tateno2017cvpr} and/or relative motion between the frames~\cite{demon, cvpr2017oral, ba-net, deeptam}. 

\subsection{Pipe reconstruction \label{sec:pipe_recon}}
\noindent
Pipe network for gas distribution or sewer is an active target of 3D reconstruction since they are often narrow or dangerous to walk in for inspection.
Instead, a robot vehicle~\cite{peter2015ijrr} or an industrial endoscope~\cite{pipe_robot, pipe_robot2} provides a safety inspection using its mounted camera(s). 
3D reconstruction from a sequence of images from the camera could also help a detailed 3D structure inspection rather than the 2D visual inspection. 
However, poor and dynamically changing lighting conditions and highly repetitive appearances due to standardized pipes make application of vision-based approaches difficult.
Several works, therefore, rely on the use of multi-domain sensors outputs, \eg, stereo camera~\cite{hansenstereo}, Inertial Measurement Unit (IMU)~\cite{esquivel2009jprs} or structured light~\cite{peter2015ijrr}, along with a fish-eye camera. 
Although rich sensors and equipment can help reconstruction, they are sometimes infeasible for very narrow structures and limited conditions, \eg, inspection of gas plumbing inside the building using an endoscope. Some works build an accurate 3D pipe model using only a monocular camera, with assumptions of camera motion and prior knowledge peculiar to the pipe. Kahi \etal~\cite{el2011robio} refine the reconstructed 3D points by cylinder fitting after BA. Zhang \etal~\cite{zhang2011wscg} and Kunzel \etal~\cite{kunzel2018wacv} rectify the images to a cylinder projection plane to triangulate points easily and regularize image illumination. 

Our system, designed for a 3D reconstruction using an industrial endoscope equipping only a monocular camera, is built based on the incremental SfM pipeline. 
Assuming a prior knowledge of the constant inner diameter of the straight pipe, we provide an accurate 3D structure by fitting 3D points to the known pipe surface. 
In contrast to the previous works limiting a camera movement to be in parallel with the pipe axis~\cite{zhang2011wscg,kunzel2018wacv}, we give no limitations about the camera path, which enables general endoscope motions during the inspection. The most relevant work of ours is \cite{el2011robio}, which detects a cylinder per reconstructed model 
and aligns the 3D points to the known pipe property after registering the fixed number of frames.
On the other hand, we detect the pipe as a general conic shape, which takes the scale-drifting errors of 3D points into account. More importantly, we search multiple pipe instances assuming the pipe network that consists of multiple pipes directed to different axes. We then incrementally refine the temporary model using a known inner diameter when each of new pipes is appeared. 
The next section describes our SfM system and its components in detail.

\begin{figure*}[ht]
    \centering
    \includegraphics[width=\textwidth]{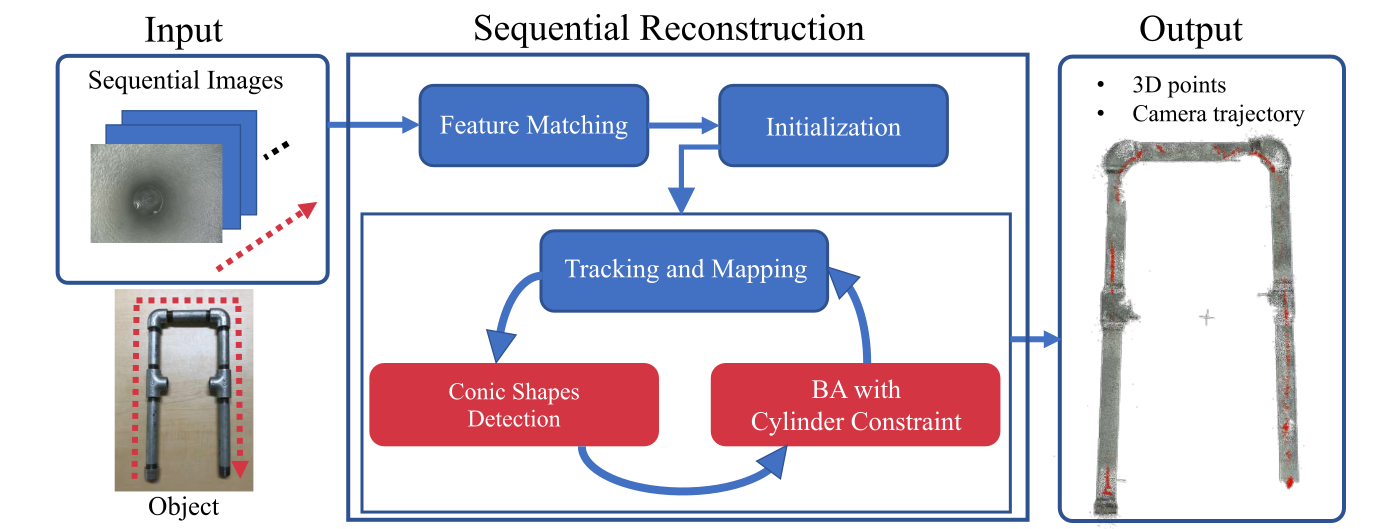}
    \caption{{\bf The overview of our SfM system for pipe reconstruction.} We design our system based on the incremental SfM pipeline for general purposes while extending it to address the particular situations of the inner pipes. 
    Our system detects multiple pipe instances as general conic shapes in the temporary model (conic shape detection) and refines the whole model as each of the detected pipes satisfies the known inner diameter (BA with cylinder constraint). }
    \label{fig:overview}
\end{figure*}

\section{Cylinder constrained SfM for pipe reconstruction}
\noindent
In this section, we describe the proposed SfM system for reconstructing a pipe network composed of multiple straight pipes using an image sequence taken by an endoscope camera. 
The main challenges in the situation are three-folded; 
{\bf 1)} Pipe networks constructed for industrial purposes are dominated by very weakly-textured (or highly repetitive) appearance (cf. Fig.~\ref{fig:objects}), and often consists of narrow and specular reflective cylindrical parts. 
In addition, pipe inspection using a endoscopic camera is often operated in a poor lighting condition, consequently suffers from local feature deprivation and inaccurate keypoints.
Vision-based system such as SfM and SLAM are critically affected by such unstable keypoint properties, resulting in an inconsistent 3D reconstruction. 
{\bf 2)} Incremental SfM iteratively registers each image in the image sequence. Therefore, the error of the temporary 3D model in each iteration is increasingly accumulated to the whole model. 
{\bf 3)} Loop closure is often infeasible during the practical visual inspection, because the camera path often has no loops due to the flexibility of the endoscope, \ie, the system cannot detect the scale-drift. 

We construct our reconstruction system (illustrated in Fig.~\ref{fig:overview}) based on the incremental SfM pipeline for general purposes (Sec.~\ref{sec:inpre}), while integrating several configuration and new processes to achieve an accurate reconstruction for the pipe network: 
Before starting reconstruction, we calibrate intrinsic parameters of the endoscope camera to mitigate the effect of lens distortion (Sec.~\ref{sec:calibration}). 
Our system finds each pipe instance from the temporary reconstructed model considering its scale-drifting (Sec.~\ref{sec:tube_detection}). 
Detected pipes are refined by our cylinder constrained BA that minimizes point distances from the cylinder surface using a known property of the tube diameter (Sec.~\ref{sec:ba_cc}). 

\subsection{Incremental SfM \label{sec:inpre}}
\noindent
In what follows, we describe an general incremental SfM pipeline for a set of sequential images. 

\para{2D feature matching. }
For each input image, SfM extracts the image local features, \eg, SIFT~\cite{sift} or improved one~\cite{dsp_sift}, to get 2D correspondences between images that are used to register the image to the model in a latter step. When the set of images are ordered by time-stamp, the matching target can be restricted within the current few frames. In our experiment, we match an image toward the current 50 frames of the sequential set. Feature correspondences are verified through an outlier rejection scheme, \eg, random sample consensus (RANSAC)~\cite{ransac,mvg}. 

\para{3D model initialization. }
Incremental SfM firstly initializes the 3D model for a selected image pair~\cite{beder2006determining,colmap}. 
We assume the input from a sequential image set, thus the selection can be easily done using time stamp, \ie, initializing the model with first two frames of the input. 
Then the initial model, composed of the 3D scene points correspond to the local feature matches and relative camera poses of the image pair, is to be constructed via two-view triangulation~\cite{beder2006determining,snavely2008scene}. 

\para{Tracking and mapping (temporary model construction). }
Once the model has been initialized, the SfM incrementally registers the input images to the model, while enriching the model by adding new 3D points correspond to 2D local feature tracks. 
In each iteration, SfM obtains local feature correspondences for a new input image (next frame) to the existing model, resulting in the set of tracked 2D observations, \ie, the keypoints seen from more than two other frames. 

Using the existing 3D points correspond to the feature tracks, SfM estimates the camera pose of the input image by solving a Perspective-n-Point (PnP) problem via P3P-RANSAC~\cite{p3p,ransac}. 
After recovering the pose of the image, the model grows adding 3D points for the newly tracked features, through the triangulation among the current frames. 

\para{Bundle adjustment. }
To stably develop the model through the incremental scheme, the system refines the temporary 3D model
after each input image registration (local bundle adjustment). 
Regarding the 3D points and camera poses of the current frames, the standard bundle adjustment~\cite{triggs1999bundle} minimizes the error of the 3D points from the corresponding 2D observations (reprojection error), which is represented as: 
\begin{equation}
   E_{rep} ({\bf X, P, K}) = \sum_{i \in {\bf P}} \sum_{j \in {\bf X}} \rho ( \| q_{ij} - \pi ({\bf P}_i, {\bf K}_i, {\bf X}_j) \|)
   \label{eq:generalba}
\end{equation}
where ${\bf X}_j$ is the $j$--th 3D scene point, $q_{ij}$ is the 2D observation of ${\bf X}_j$ from the $i$--th view ${\bf P}_i$, $\pi ({\bf P}_i, {\bf K}_i, {\bf X}_j)$ is a function that projects scene points to the image plane, and $\rho$ is the robust function, \eg, Cauchy function. 

The system also runs another refinement process after several iterations (global bundle adjustment) that maintains the consistency of the whole model. In this time, bundle adjustment also minimizes Eq.~(\ref{eq:generalba}) but in regard to all 3D points and frames registered to the model. 

\subsection{Endoscope camera calibration \label{sec:calibration}} 
\begin{figure}[t]
    \begin{minipage}{0.64\linewidth}
        \centering
        \includegraphics[width=\linewidth]{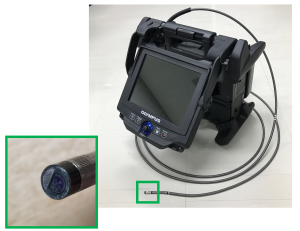}
    \end{minipage}%
    \begin{minipage}{0.34\linewidth}
        \centering
        \includegraphics[width=\linewidth]{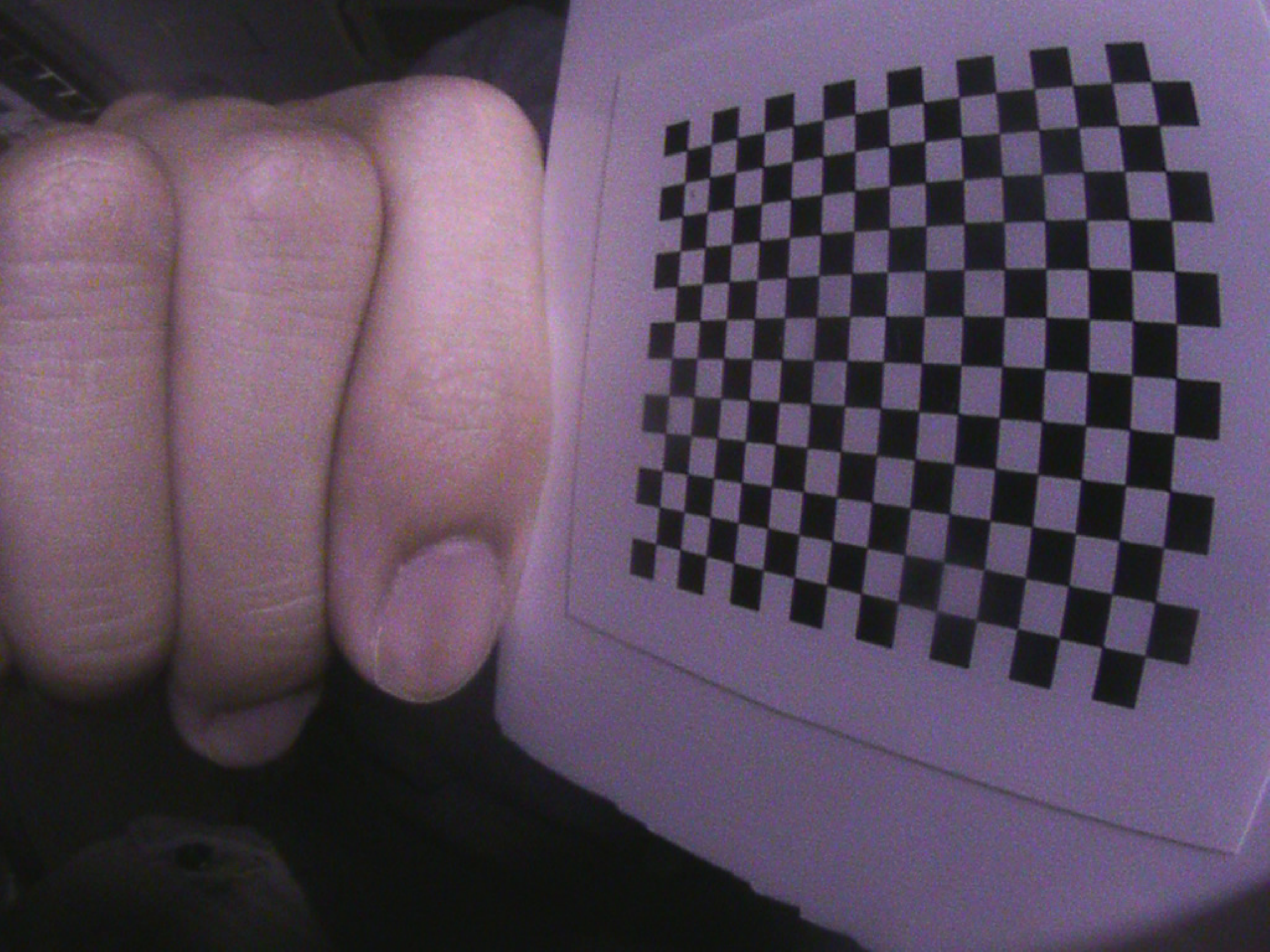}
        \par \medskip \vfill
        \includegraphics[width=\linewidth]{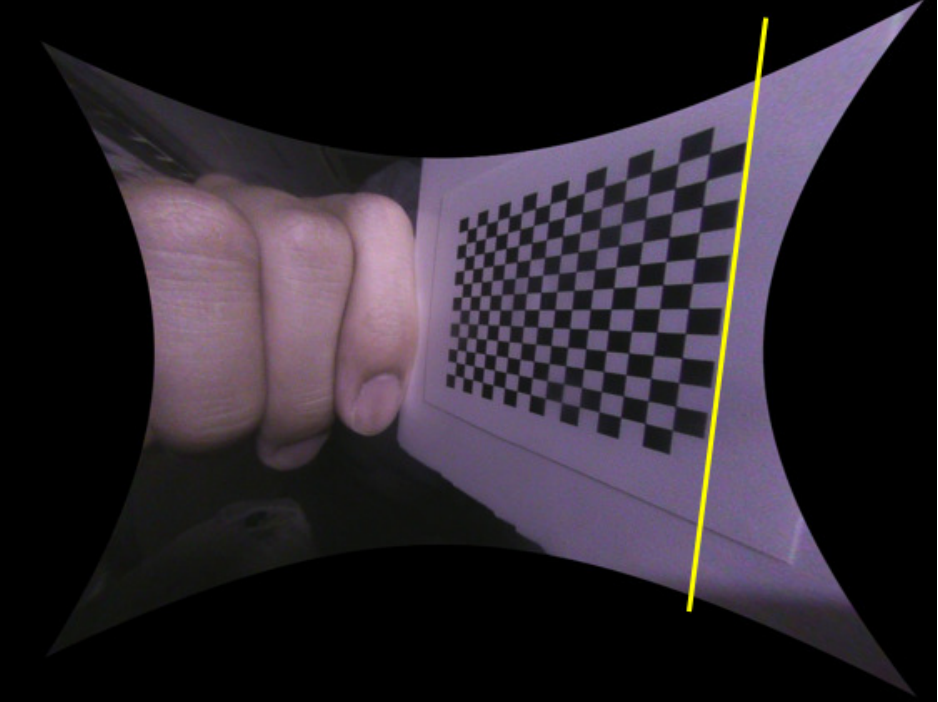}
    \end{minipage}
    \caption{{\bf Industrial endoscope.} Left: The appearance of the industrial endoscope. We use an Olympus IPLEX NX with the AT120D/NF-IV96N optical adapter and the IV9635N scope. Right: A sample image capturing checkerboard pattern (top) and rectified by fish-eye camera model (bottom). \label{fig:hardware}}
\end{figure}

\noindent
An accurate intrinsic parameter of the camera, that makes a relation between an image point $[u, v]^{\mathrm{T}}$ and a 3D point (normalized image coordinate) $[x, y, z]^{\mathrm{T}}$, is required to achieve a solid 3D reconstruction, \eg, for an accurate conversion $\pi$ in Eq.~(\ref{eq:generalba}).  
We focus on the use of a standard industrial endoscope (Fig.~\ref{fig:hardware}, Tab.~\ref{tab:spec}) which has a wide FoV camera for efficient visual inspection. 
To deal with the image distortion comes from the wide FoV configuration, we 
rectify the image coordinates of the keypoints assuming a fish-eye model~\cite{fisheye}. 
By the camera model, the relation between the image coordinate and the 3D point is formulated as: 
\begin{equation}
    \centering
    \left[
    \begin{array}{r}
    u \\
    v
    \end{array}
    \right] 
    = 
    \left[
        \begin{array}{r}
        f_x(\theta +  k_1 \theta^3 + k_2 \theta^5)cos\phi + u_0 \\
        f_y(\theta +  k_1 \theta^3 + k_2 \theta^5)sin\phi + v_0
        \end{array}
    \right]
\end{equation}
where $\theta$ is the 3D angle formed by the camera optical axis and the ray going from camera center to the 3D point, 
and $\phi$ is the polar angle of normalized image coordinate, respectively. 
\begin{equation}
    \centering
    \theta = \arctan\left(\sqrt{\frac{x^2+y^2}{z^2}}\right), \phi = \arctan\left(\frac{y}{x}\right)
\end{equation}
Camera intrinsic parameter consists of: 
focal length with respect to horizontal and vertical axis $(f_x, f_y)$, 
distortion parameters $(k_1, k_2)$, 
and image principal point $(u_0, v_0)$. 
Before starting the reconstruction, we initialize the parameters by an offline calibration. To achieve an accurate calibration, we use a checkerboard pattern with a 2mm size of each square (Fig.~\ref{fig:hardware}) and take pictures from multiple views. The parameters are found by minimizing the sum of squared reprojection errors of the grid points~\cite{opencv_library}. 
We also update the parameters during the reconstruction via bundle adjustment (Sec.~\ref{sec:ba_cc}). 

{\tabcolsep=3pt
\begin{table}[t]
  \centering
  \caption{The detailed specification of the industrial endoscope \label{tab:spec}}
  {\footnotesize
  \begin{tabular}{ccccc}
    Resolution & \specialcell{Field of \\ view} & \specialcell{Scope \\ diameter} & \specialcell{Depth of \\ field} & Illumination \\ \hline
    \rule{0pt}{2ex} $1024 \times 768$[px]  & $120^{\circ}$ & 6.0 mm & 7 to 300 mm & laser diode
  \end{tabular}
  }
\end{table}
}

\subsection{Incremental conic shapes detection \label{sec:tube_detection}}
\noindent
In spite of the effort of the precise camera calibration, general incremental SfM pipeline can often cause a significant 3D model distortion on a pipe inner situation. 
We address the errors assuming the known properties of a pipe network, \ie, if we can detect the pipe instances that is a component of the pipe network, the model can be refined by fitting 3D points to the known properties. 
Instead of detecting the cylinders for the whole 3D model, which would requires substantial conversion of the model because of accumulated model distortion, we search the pipe instance from the temporary 3D points during the incremental pipeline.

Also, we observed even a small intrinsic calibration error can cause a large scale-drifting of the temporary model (Fig.~\ref{fig:color_results}), which can limit the standard cylinder detection for the 3D point cloud. 
Therefore, we search the pipe by fitting a general conic shape~\cite{Lopez2014automatic} to the 3D points. 

In each temporary model constructed during the incremental SfM process, we assume a 3D point in the homogeneous coordinate system $\bm{X}=[x,y,z,1]^{\mathrm{T}}$ represents the surface of a cone if it satisfies:
\begin{equation}
    \centering
    \bm{X}^{\mathrm{T}} {\bf C}\bm{X} = 0
    \label{eq:cone}
\end{equation}
${\bf C}$ is a symmetric matrix, which can be decomposed as: 
\begin{equation}
    {\bf C} = 
    \left[
        \begin{array}{cc}
        R^{\mathrm{T}}{\bf D}R & R^{\mathrm{T}}{\bf D}t \\
        t^{\mathrm{T}}{\bf D}R & t^{\mathrm{T}}{\bf D}t
        \end{array}
    \right], 
    {\bf D} = {\mathop{\rm diag}\nolimits}(-c^2, -c^2, 1)
    \label{eq:conedecomp}
\end{equation}
where $c$ is the constant parameter representing the slope of the cone, and $[R, t]$ is the 3D rotation and translation that represents a coordinate transformation which aligns z--axis to the major axis of the cone. 
Eq.~(\ref{eq:cone}) can also be rewritten by: 
\begin{equation}
  vech^{\mathrm{T}} (\bm{X}\bm{X}^{\mathrm{T}}) vech({\bf C}) = 0
  \label{eq:cone_vech}
\end{equation}
$vech()$ is the half vectorization transformation of a symmetric matrix that is obtained by vectorizing the lower triangular part of the matrix. 
Minimal solution for Eq.~(\ref{eq:cone_vech}) is therefore given by nine points. 

We incrementally find multiple pipe instances by fitting cone to the newly produced 3D points. In each iteration of the incremental SfM, we search a cone in the temporary 3D model via RANSAC~\cite{ransac}. 
The registered images are then labeled as it belongs to the pipe or not, based on the number and the ratio of inliers that support the cone model. 
Once a pipe instance is detected, the detector searches a new conic shape from the 3D points seen by current frames that do not belong to any existing instances. 
All cone parameters are then refined by local hypotheses refinement~\cite{chum2003locally} using 3D points observed by the labeled images, which achieves optimal cone fitting for 3D points. 

\subsection{Bundle adjustment with cylinder constraint \label{sec:ba_cc}}
\noindent
Bundle adjustment in the general incremental SfM pipeline refines the current model by minimizing reprojection error represented by Eq.~(\ref{eq:generalba}). 
If the conic shape detector finds the straight pipes, we can also compute the error of the 3D points with respect to the prior knowledge of the pipe properties, which is formulated as: 
\begin{equation}
  E ({\bf X, P, K, C}) = E_{rep} ({\bf X, P, K}) + \alpha E_{cyl} ({\bf X, C})
  \label{eq:cost_ba}
\end{equation}
where the first term $E_{rep} ({\bf X, P, K})$ is the reprojection error term which is equal to Eq.~(\ref{eq:generalba}), and the second term $E_{cyl} ({\bf X, C})$ is our new cylinder constraint term. 
${\bf X, P, K,}$ and ${\bf C}$ are the variables that indicate the sets of 3D points, the camera poses of the registered images, the camera intrinsic parameters, and the detected cones parameters, respectively. $\alpha$ is a constant scalar which controls the weights of two competing error terms. 

Our cylindrical constraint punishes the distance of the 3D points $\bm{X}_j$ from the cylinder surface around the major axis. 
Assuming the uniformly distributed accumulated error, the axis of the cylinder can be approximated by detected axis of the cone $[R_i, t_i]$, which is given by the decomposition of cone parameters ${\bf C}_i$ according to Eq.~(\ref{eq:conedecomp}). 
$E_{cyl}$ is therefore formulated by: 
\begin{equation}
   E_{cyl} ({\bf X, C};r) = \sum_{i} \sum_{j} \rho (\| r - d(\bm{X}_j,[R_i, t_i])\| )
\end{equation}
where $d(\bm{X}_j,[R_i, t_i])$ is the distance of the 3D point $\bm{X}_j$ from the major axis of the cylinder, 
and $r$ is the known inner diameter of the pipe. 
$\rho$ is the Cauchy function used as the robust function. 

Our incremental SfM system includes two types of bundle adjustment, local BA and global BA. 
After registering each input image, our system performs local BA that refines camera poses of current frames, intrinsic parameters, and 3D points, by minimizing Eq.~(\ref{eq:cost_ba}). 
When the model grows by a certain percentage or a new pipe instance is detected, the system runs global BA that optimizes all model parameters including the straight pipe parameters. 
For completeness and fastness, we refine cameras intrinsic only after detecting first straight pipe. 
Please notice that our newly proposed error term does not give any constraint on camera motion, unlike the previous works~\cite{el2011robio,zhang2011wscg,kunzel2018wacv}. Also notice that our SfM system updates each temporary model, thus produces a substantial different results from the one constructed via standard SfM at the end. As shown in the next section, this property enables us to obtain further complete and accurate 3D model (cf. Fig.~\ref{fig:qualitative}).

{\tabcolsep=1pt
\begin{figure}[t]
    \centering
    \begin{tabular}{cccc}
    \rot{\rlap{~ \footnotesize Network A}} 
    &
    \includegraphics[width=0.3\linewidth]{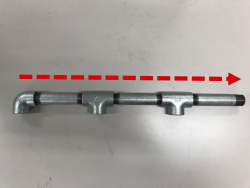}
    &
    \includegraphics[width=0.3\linewidth]{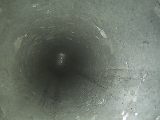}
    &
    \includegraphics[width=0.3\linewidth]{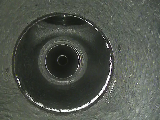}
    \\
    \rot{\rlap{~ \footnotesize Network B}}
    &
    \includegraphics[width=0.3\linewidth]{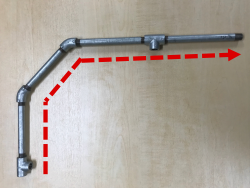}
    &
    \includegraphics[width=0.3\linewidth]{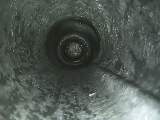}
    &
    \includegraphics[width=0.3\linewidth]{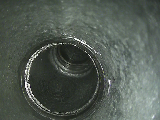}
    \\
    \rot{\rlap{~ \footnotesize Network C}}
    &
    \includegraphics[width=0.3\linewidth]{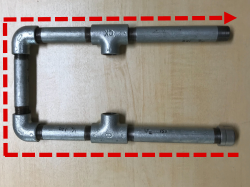}
    &
    \includegraphics[width=0.3\linewidth]{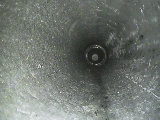}
    &
    \includegraphics[width=0.3\linewidth]{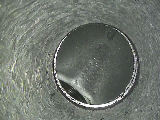}
    \\
    \rot{\rlap{~ \footnotesize Network D}}
    &
    \includegraphics[height=0.225\linewidth]{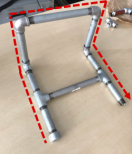}
    &
    \includegraphics[width=0.3\linewidth]{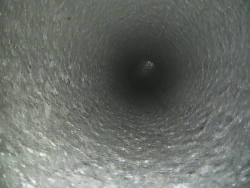}
    &
    \includegraphics[width=0.3\linewidth]{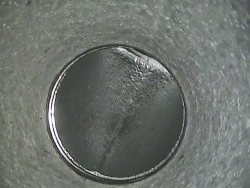}
    \end{tabular}
    
    \caption{{\bf Datasets.} Left: the target pipe network and camera path (red arrow) of each sequence. Middle and right: sample images of each sequence (middle: at a straight pipe, right: at an elbow).\label{fig:objects}}
\end{figure}
}
{\tabcolsep=11pt
\begin{table}[t]
  \centering
  \caption{{\bf The properties of each dataset.} We setup four different types of pipe networks and capture videos using the standard industrial endoscope as shown in Fig.~\ref{fig:hardware}. Note that the material of all pipes is steel.\label{tab:properties}}
  \begin{tabular}{l|cccc}
    Properties & A & B & C & D \\ \hline
    Inner diameter [mm] & 16.1 & 8.0 & 16.1 & 16.1 \\ 
    Video times [sec] & 61 & 183 & 367 & 442 \\
    Centering device & Yes & No & No & No \\ \hline
    \# Straight pipe & 3 & 4 & 5 & 7 \\
    \# Tee & 2 & 2 & 2 & 2 \\
    \# Elbow & 1 & 2 & 2 & 4 \\
    Elbows angle (max) & $0^{\circ}$ & $45^{\circ}$ & $90^{\circ}$ & $90^{\circ}$ \\
  \end{tabular}
\end{table}
}

\section{Experiments}
\noindent
In this section, we describe the performance of our incremental SfM pipeline designed for a pipe network reconstruction. We evaluate our method on several image sequences capturing the inners of pipe networks, which consist of multiple industrial pipes and indicate the practical scenarios of industrial visual inspection. 
We firstly demonstrate our new constrained BA in an incremental SfM system effectively deals with error accumulation during the reconstruction (Sec.~\ref{sec:ba_impact}). We next compare our method with several vision-based reconstruction systems on our dataset (Sec.~\ref{sec:comparison}). 

\para{Environment. }
Fig.~\ref{fig:objects} shows the four types of pipe networks we set up for evaluation. Detailed properties are described in Tab.~\ref{tab:properties}. All scenes consist of 1$\sim$5 straight pipes, tees, and elbows, constructed by industrial steel parts. 
We collect $60$fps image sequences of those networks using an industrial endoscope (Fig.~\ref{fig:hardware}). Detailed specifications of the endoscope are summarized in Tab.~\ref{tab:spec}. The camera moves backward inside the networks following the red arrows in Fig.~\ref{fig:objects}, which is the practical manner of the pipe inspection due to the physical limitation of the endoscope. 

The network A consists of three straight pipes in the same direction. 
In this simple structure, we optionally attach an guide head device for the endoscope that roughly forces the camera to be in center of the pipe. 
Note that this setting fairly makes the camera path to be stable and makes the reconstruction easier, but is sometimes infeasible, because the guide head does not support significant direction changes, \eg, curved camera path at an elbow. We do not use this attachment for network B, C, and D, to depict more general situations of the pipe network inspection. 
Network B has a narrower inner diameter (8.0mm) than others (16.1mm) which leads to a more severe appearance changes and occlusions at the elbow parts. 
Network D consists of the maximum number of pipes and elbows according to the flexibility of the endoscope, connected three-dimensionally with independent orientation. 

\para{Evaluation metric. }
To evaluate the accuracy of 3D models, we compute the reconstruction error as the difference from the prescribed inner diameter of each straight pipe.
RMSE of radius rate is determined as:
\begin{equation}
  RMSE = \sqrt{\frac{1}{n} \sum^n_i \left(\frac{r_i - r}{r} \right)^2}
  \label{eq:rmse}
\end{equation}
where $r_i$ is the radius of inlier points in the straight cylinders estimated by our SfM, and $r$ is the prescribed radius value. 
For other methods that originally do not detect any pipe, we additionally detect cylindrical parts and scale the model for evaluation, after the whole reconstruction process. 
Specifically, we fit the multiple cones to the reconstructed model via sequential RANSAC while giving the number of pipe parts. The model is then scaled as approximating the diameter of the pipe by the average of points distances from the cone axis. 

\para{Implementation. }
We construct our system based on an incremental SfM system implemented by COLMAP~\cite{colmap}, a widely known reconstruction tool. After constructing temporary 3D model for each 30 frames of the input, the system searches and refines the pipe instances of the model as described in Sec.~\ref{sec:tube_detection}. Once a cylinder is detected, the system replace bundle adjustment process in each iteration by our cylinder constrained BA (Sec.~\ref{sec:ba_cc}). We assume the pipe inner diameter of each Network is constant and known (cf. Tab.~\ref{tab:properties}). We experimentally set the parameter $\alpha$ in Eq.~(\ref{eq:cost_ba}) by 10. 

{\tabcolsep=7pt
\begin{table}[t]
  \footnotesize
  \centering
  \caption{{\bf Quantitative results.} Evaluation of 3D reconstruction results by our datasets. The values are the radius error rate (RMSE) (Eq.~(\ref{eq:rmse})). \label{tab:performance}}
  \begin{tabular}{l|cccc}
    Method & A & B & C & D \\ \hline
    ORB-SLAM & 0.1916 & 0.5428 & 0.3331 & ${\bf 0.0958}$ \\
    COLMAP & 0.1615 & 0.3291 & 0.3055 & 0.2670 \\
    {\bf Ours} & $\bf{0.1375}$ & $\bf{0.2719}$ & $\bf{0.1560}$ & 0.1034 \\
  \end{tabular}
\end{table}
}

\renewcommand{\thiswidth}{0.19\linewidth}
{\tabcolsep=1pt
\begin{figure}[t]
    \centering
     \begin{tabular}{ccccc}
          \includegraphics[height=0.40\linewidth]{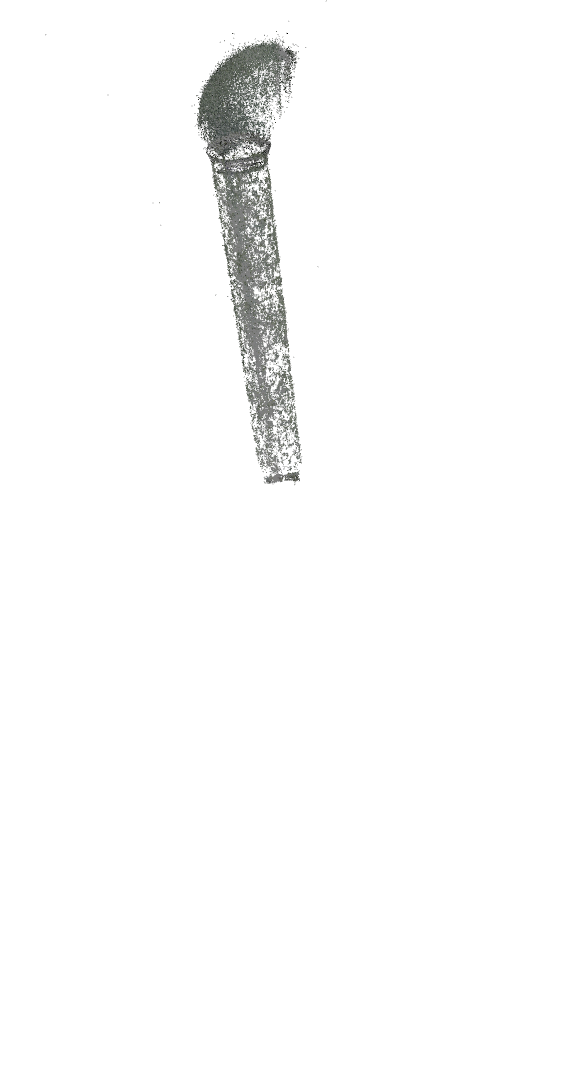} &
          \includegraphics[height=0.40\linewidth]{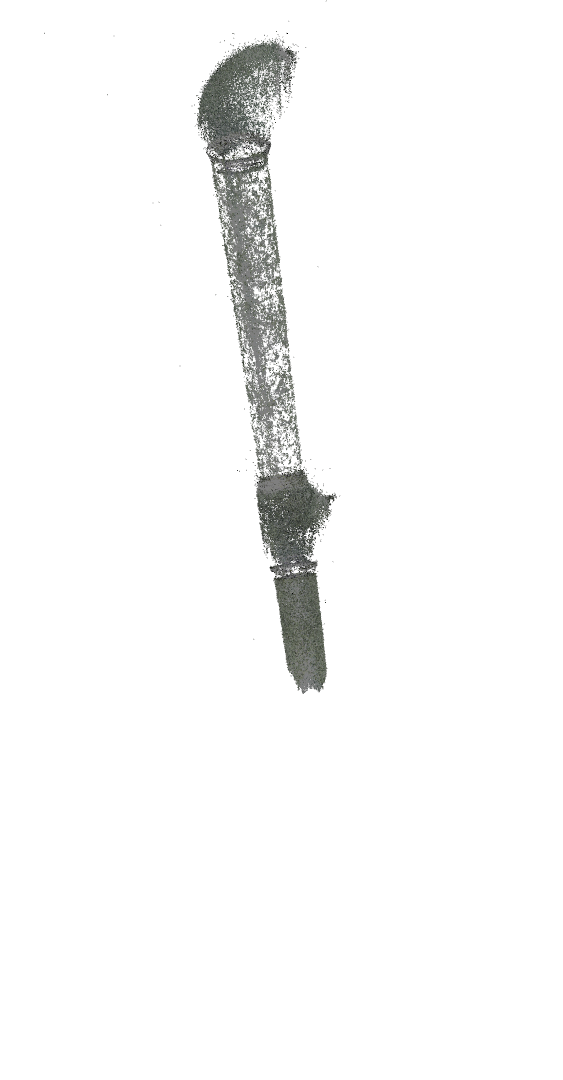} &
          \includegraphics[height=0.40\linewidth]{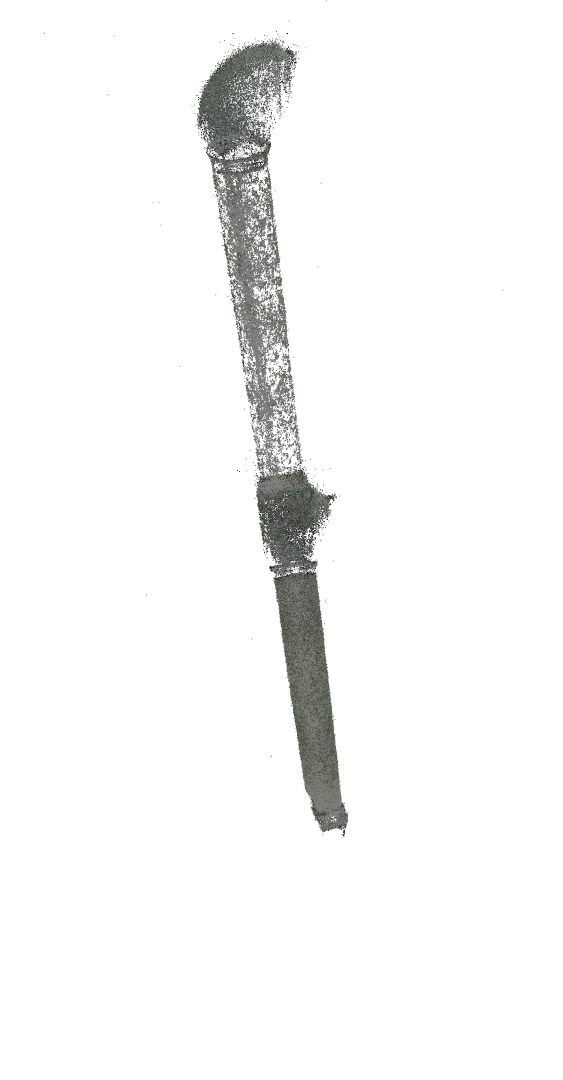} &
          \includegraphics[height=0.40\linewidth]{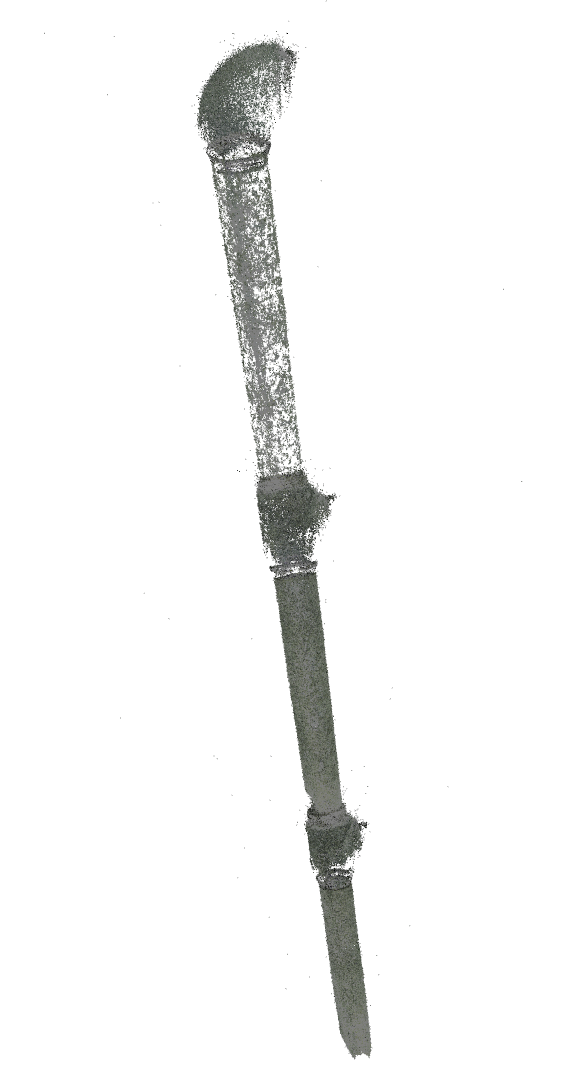} &
          \\
          \includegraphics[height=0.40\linewidth]{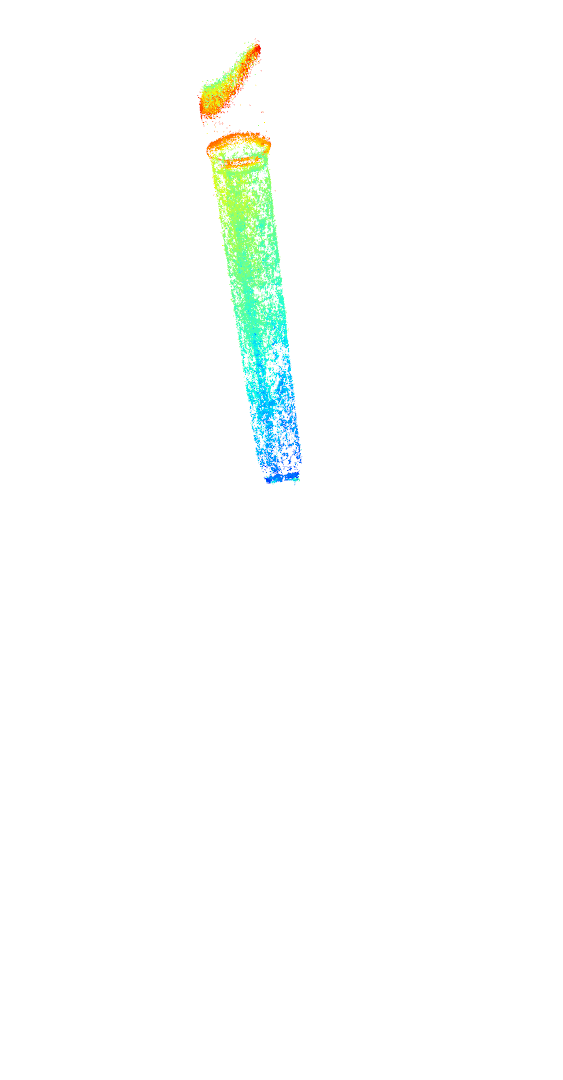} &
          \includegraphics[height=0.40\linewidth]{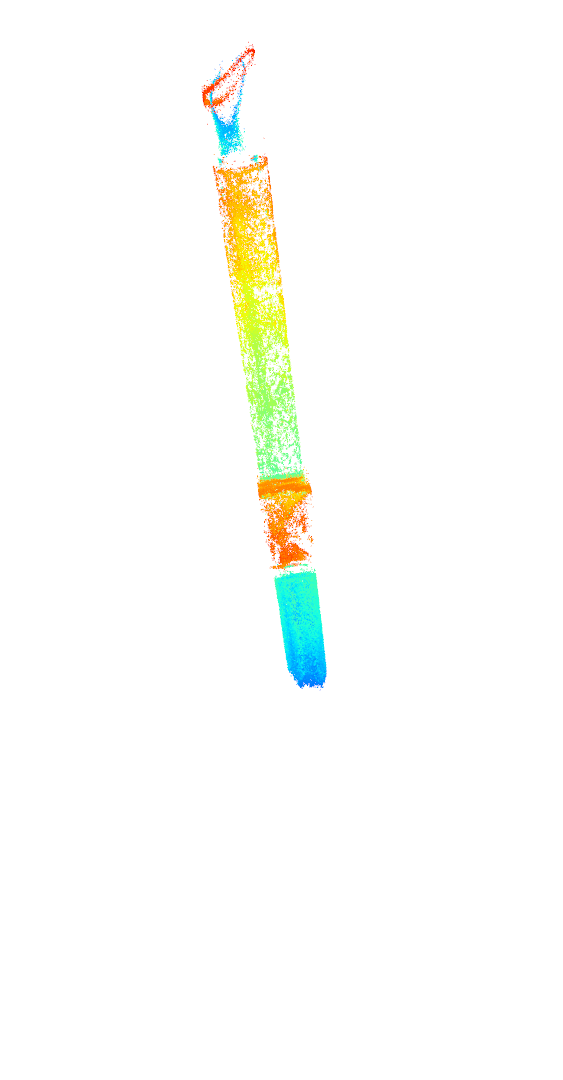} &
          \includegraphics[height=0.40\linewidth]{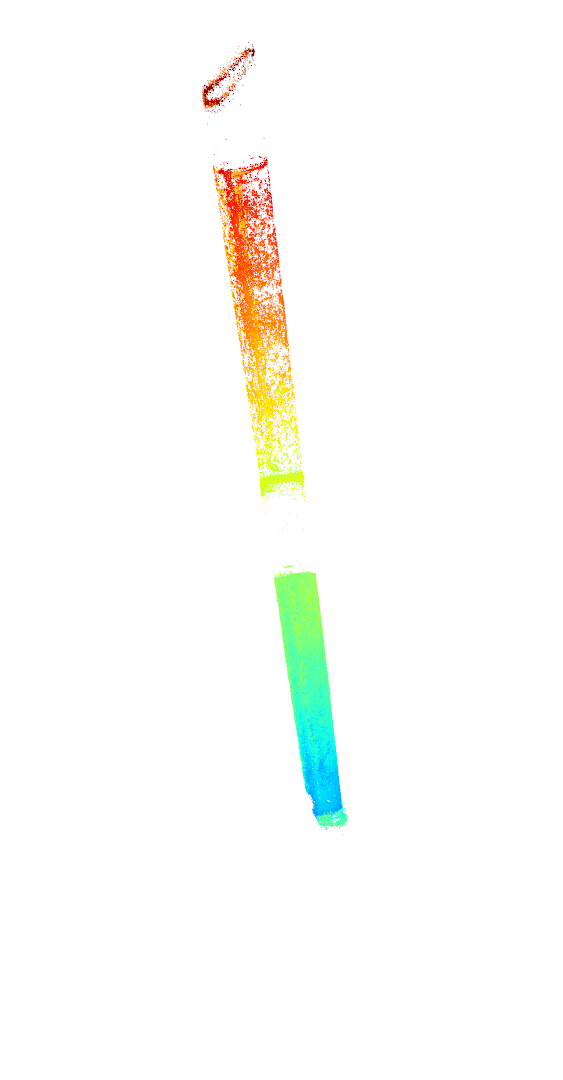} &
          \includegraphics[height=0.40\linewidth]{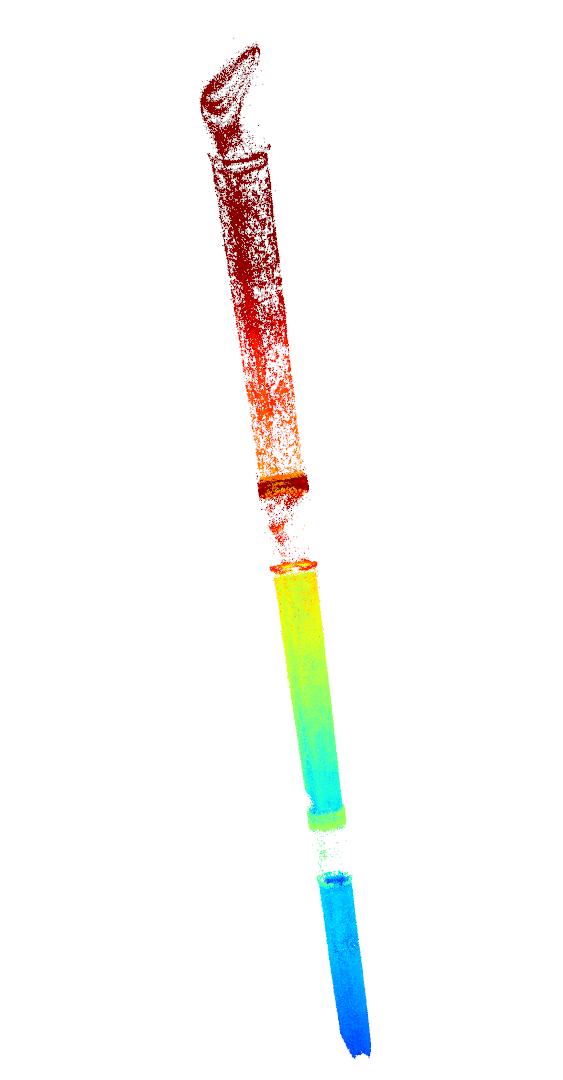} & 
          \includegraphics[height=0.40\linewidth]{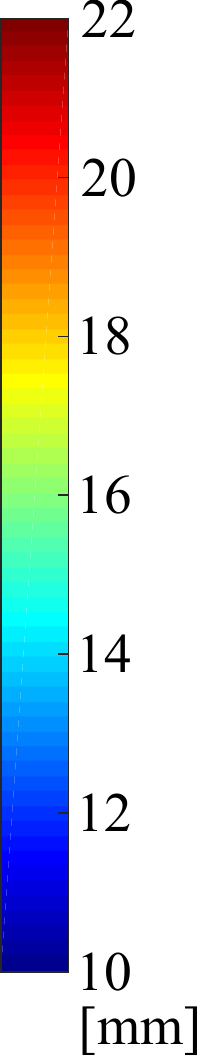} \\[-5pt]
          \multicolumn{5}{c}{{\footnotesize (a) 3D points obtained via COLMAP. }} \\[5pt]
          \includegraphics[height=0.40\linewidth]{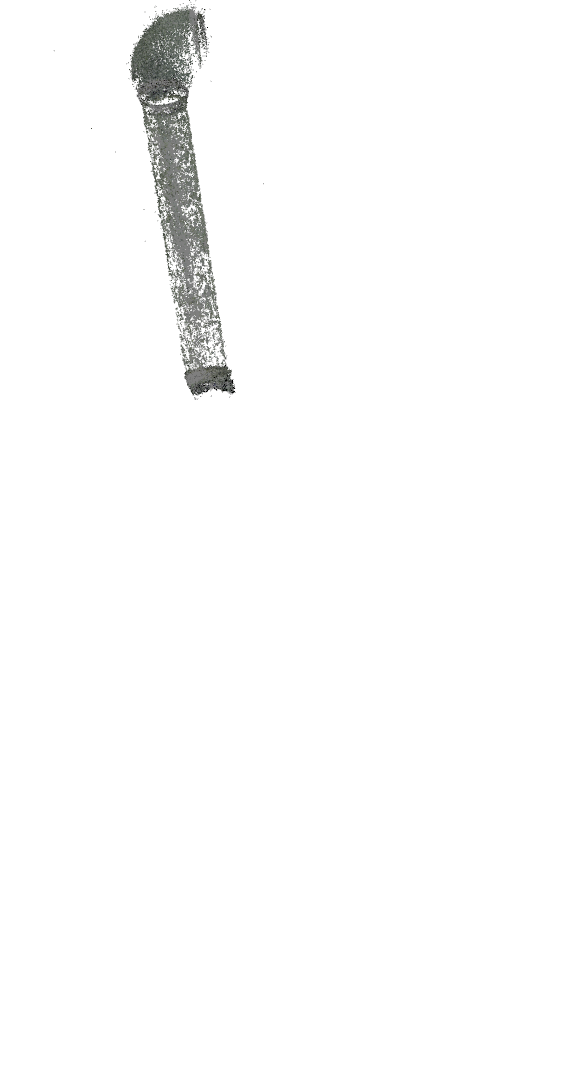} &
          \includegraphics[height=0.40\linewidth]{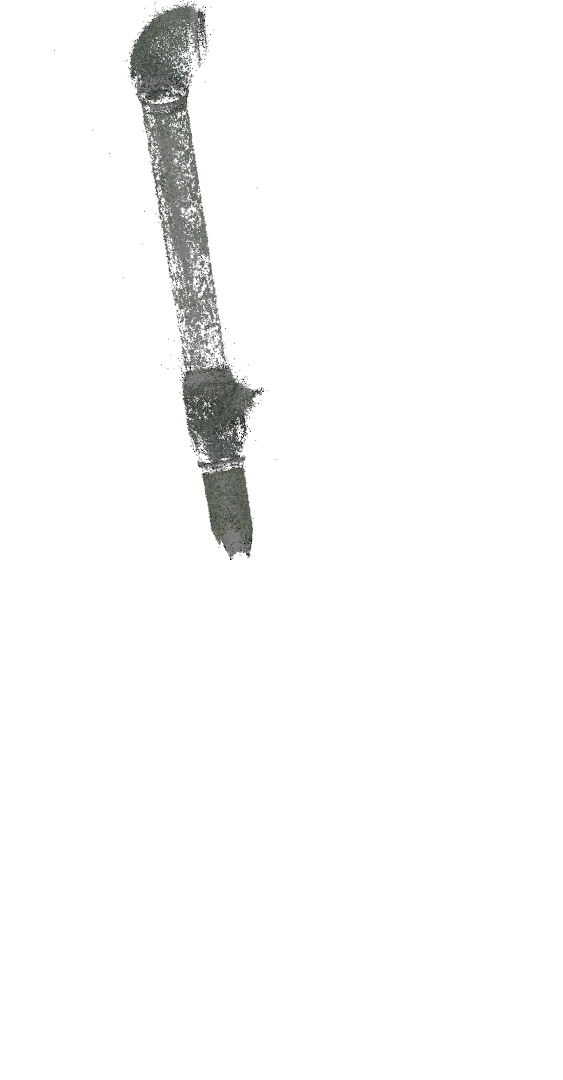} &
          \includegraphics[height=0.40\linewidth]{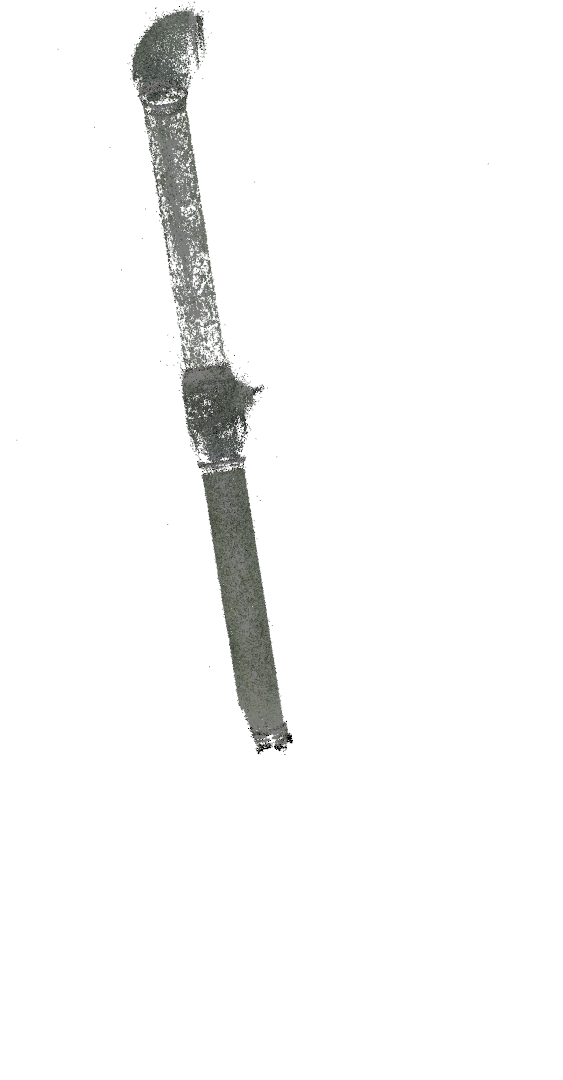} &
          \includegraphics[height=0.40\linewidth]{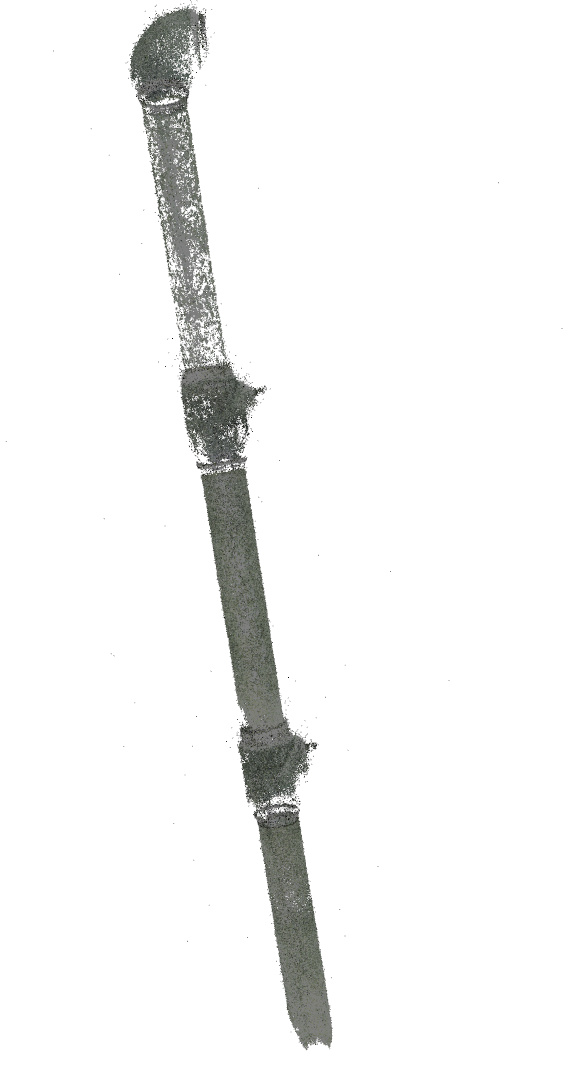} &
          \\
          \includegraphics[height=0.40\linewidth]{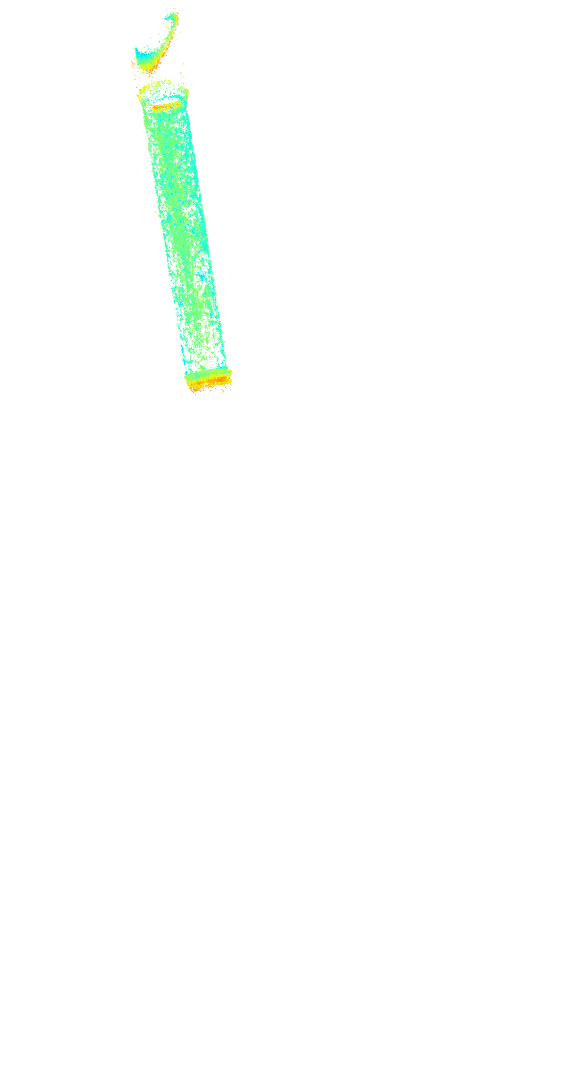} &
          \includegraphics[height=0.40\linewidth]{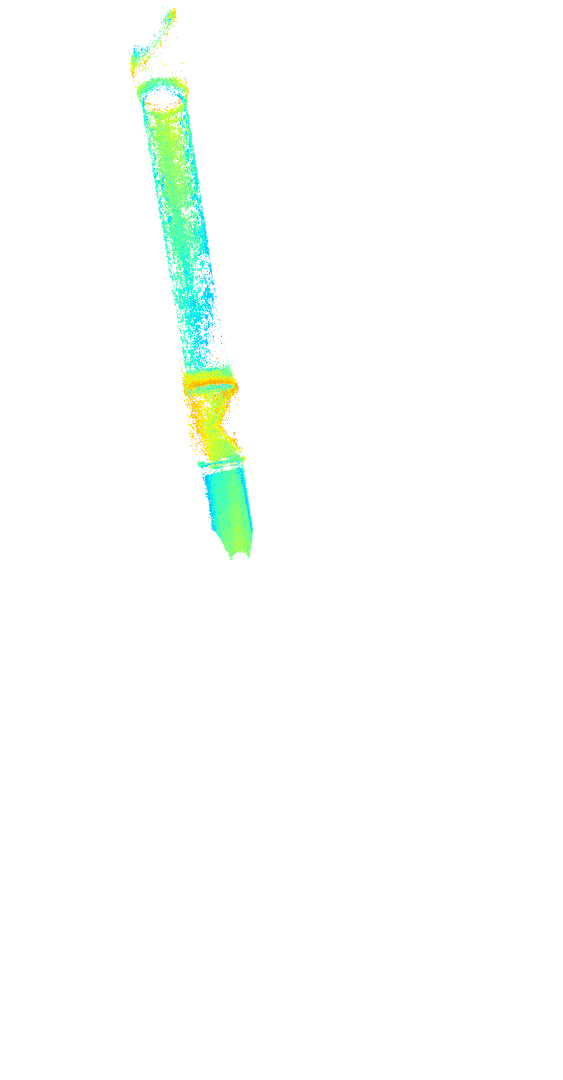} &
          \includegraphics[height=0.40\linewidth]{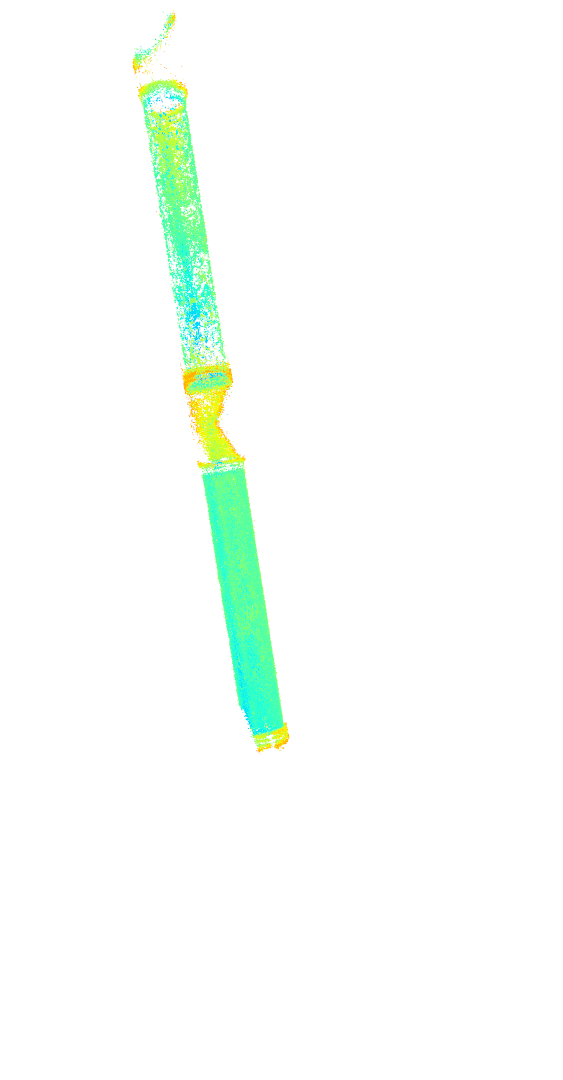} &
          \includegraphics[height=0.40\linewidth]{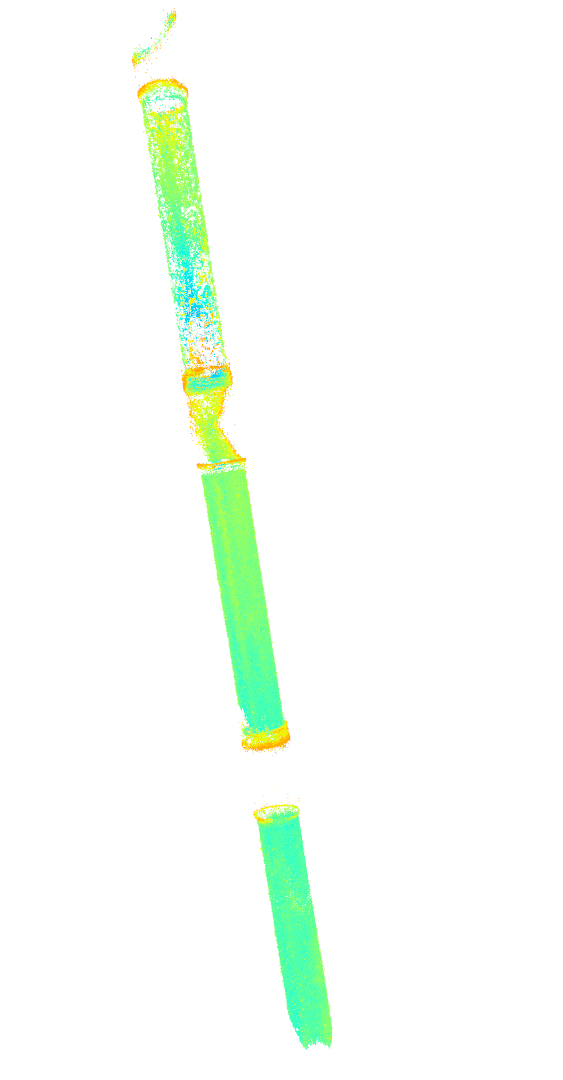} & 
          \includegraphics[height=0.40\linewidth]{colorbar.pdf} \\[-5pt]
          \multicolumn{5}{c}{{\footnotesize (b) 3D points obtained via proposed method. }}
      \end{tabular}
      \caption{{\bf Progress of the temporary model during the incremental SfM reconstruction.}
      3D points show how the temporary model obtained via (a) COLMAP and (b) our incremental SfM pipeline grow up when each of the new images is registered. From left to right, each column roughly associates the temporary model updated by the end of the first pipe, the middle of the second pipe, the end of the second pipe, and the end of the pipe network, respectively. 
      Color-coded 3D points indicate the distance of each 3D point from the major axis of the pipe, regarding the true diameter of the pipes as 16.1mm. 
    }
      \label{fig:color_results}
\end{figure}
}

\subsection{Impact of cylinder constrained BA \label{sec:ba_impact}}
\noindent
To demonstrate the impact of our cylinder constrained BA using the known pipe property, 
we evaluate our method on Network A, the simplest situation on which all pipes direct to a common axis. For a comparison, we also run an incremental SfM for general purposes (COLMAP)~\cite{colmap}, which refines the model by minimizing Eq.~(\ref{eq:generalba}), on the same sequence. 
Fig.~\ref{fig:qualitative}~(c,d) shows the appearances of the models obtained by proposed SfM and COLMAP. 
Both methods succeed to recover the whole pipe network but proposed provides more accurate model (cf. Tab.~\ref{tab:performance}). 
Fig.~\ref{fig:color_results} shows the 3D points of the temporary models obtained during the reconstruction, via COLMAP (a) and proposed (b). 
During the reconstruction, COLMAP increasingly produces large errors regarding the known inner diameter, due to the errors of intrinsic parameters and erroneous estimations of the camera motion. 
On the other hand, proposed method iteratively detects and refines each of the pipe instances, resulting in more accurate 3D model regarding the constant inner diameter. 

\subsection{Reconstruction of multiple pipes network \label{sec:comparison}}
\noindent
Next we compare our SfM system with several other reconstruction systems on the pipe networks consist of multiple straight pipes. 

\para{Comparisons. }
We compare our method to the state-of-the-art methods in each of three approaches described in Sec.~\ref{sec:related_work}, COLMAP~\cite{colmap} for SfM, ORB-SLAM~\cite{orbslam} for feature-based SLAM, and DSO~\cite{dso} for direct SLAM, respectively. For each comparison, we use the implementation provided by the authors. For a fair comparison, we use a calibrated fish-eye camera model (Sec.~\ref{sec:calibration}) for all methods\footnote{Since the original ORB-SLAM implementation only accepts a perspective model, we extend it for the fish-eye model.}. Note that we use DSO without photometric calibration since it is difficult to collect calibration data for industrial endoscope because of a built-in light source. While we use input sequence as $60$ fps for real-time methods (ORB-SLAM and DSO), we use $5$ fps sequence for offline methods (ours and COLMAP). We do not compare our methods with several works designed for a single pipe~\cite{peter2015ijrr,el2011robio, zhang2011wscg, kunzel2018wacv} because they do not provide their original implementations. But we believe adapting these works for each reconstructed pipe as batch-like process does not much improve reconstruction since they highly depend on the quality of the initial model, which is often largely distorted as shown in Sec.~\ref{sec:ba_impact}. 

Tab.~\ref{tab:performance} shows the quantitative evaluation of 3D reconstruction results of COLMAP~\cite{colmap}, ORB-SLAM~\cite{orbslam}, and ours. Our method outperforms the baseline COLMAP on all scenes, and the margin is remarkable in network C. For network D, ORB-SLAM gives the best RMSE score, but it reconstructs only two pipes in the scene. 
Fig.~\ref{fig:qualitative} shows the qualitative results of each method in each scene.
While ORB-SLAM and DSO can reconstruct in real-time, their reconstruction results are not as accurate as offline methods like ours and COLMAP, especially in complex scenes as network C and D. 
ORB-SLAM reconstructs all the images in A and B, but fail to track the image sequences in C and D. This failure of tracking occurs because the method assumes a constant velocity motion to track the camera that is often not functional for an endoscopic cameras motion, \eg, significant view changes in an elbow. 
In contrast, the proposed method reconstructs the whole target for all sequences, while preserving a stable diameter. 

\para{Limitation. }
To address the difficulties raised in the pipe inner situation, our system applies a prior knowledge (\ie constant inner diameter) to the existing 3D points in the model, rather applying before or during the 3D mapping, \eg, feature matching guided by known properties, or 3D points triangulation constrained within the known pipe surface. This strategy, 
however, could result in insufficient model reconstruction when the system cannot find sufficient matches. The problem is caused when the pipe network includes pipes which have especially severe properties, \eg, material which has a smooth nature. Potential approach to make the system robust to such severe conditions is that obtaining matches in dense manner~\cite{tola2009daisy,aji_sfm} attempting to get pixel-wise precise matches and offering outlier rejection scheme guided by pipe properties. 

Another future work is to determine a proper parameter $\alpha$ in Eq.~\ref{eq:cost_ba}, which balances the temporal property and the prior information, also regarding the demand of the model quality. 

\renewcommand{\thisheight}{0.37\columnwidth}
{\tabcolsep=3pt
\begin{figure*}[t]
    \centering
     \begin{tabular}{cc|cccc}
        \rot{\rlap{~ \footnotesize \hspace{15pt} Network A}} &
        \includegraphics[height=\thisheight]{sequenceA_appearance_ds.png}
        &
            \includegraphics[clip, height=\thisheight]
            {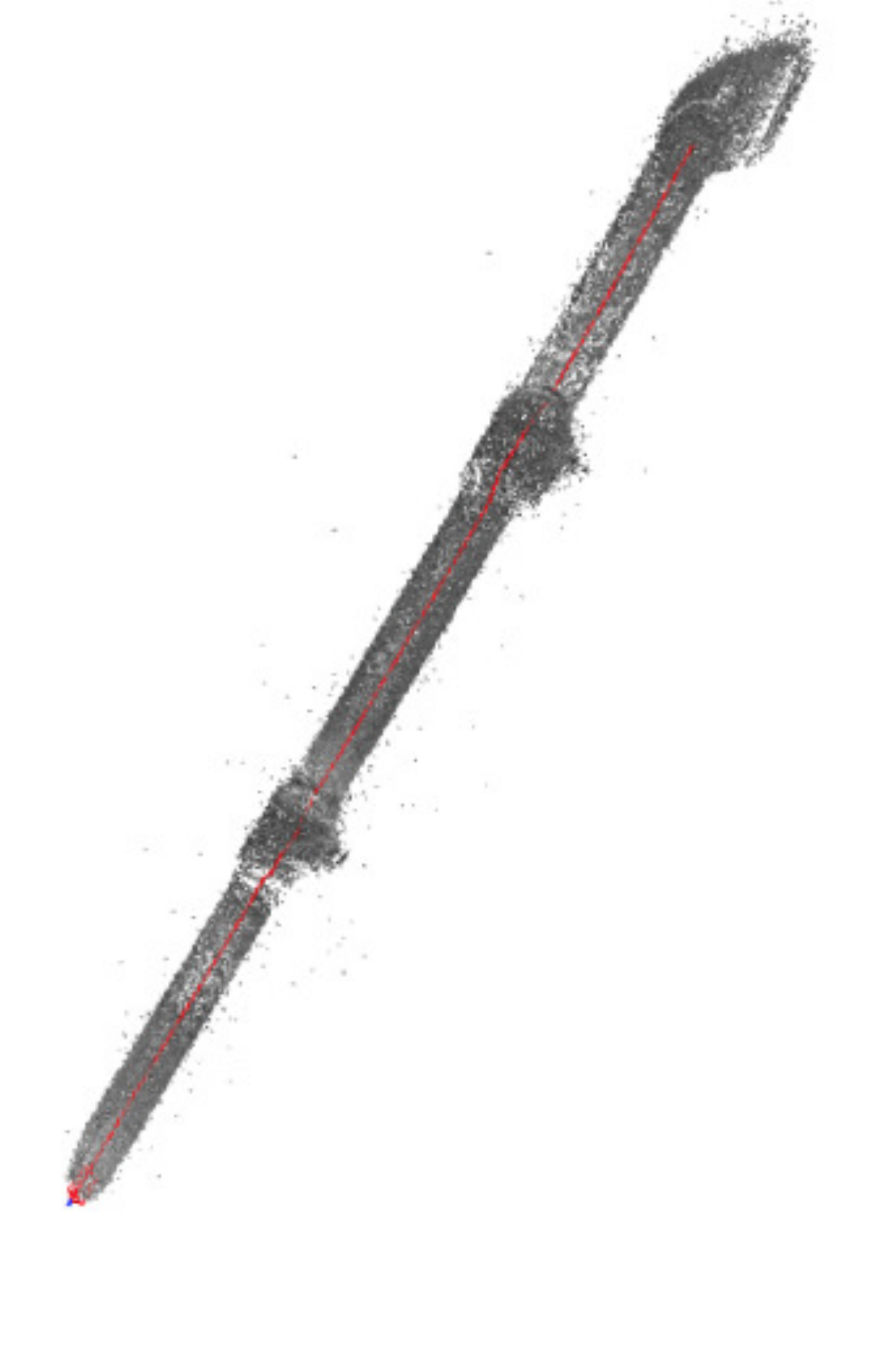}
         &
            \includegraphics[clip, height=\thisheight]
            {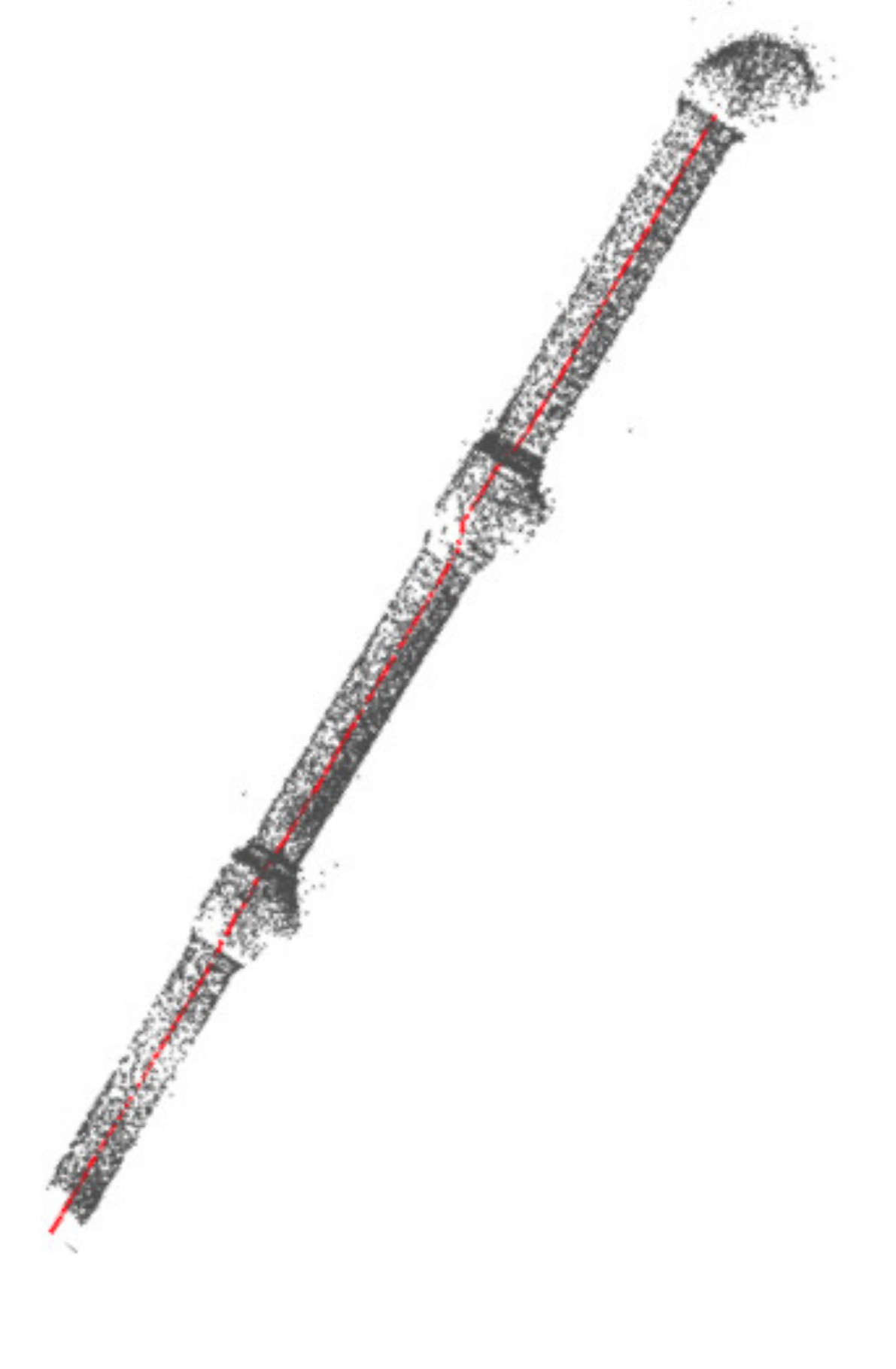}
         &
          \includegraphics[clip, height=\thisheight]
            {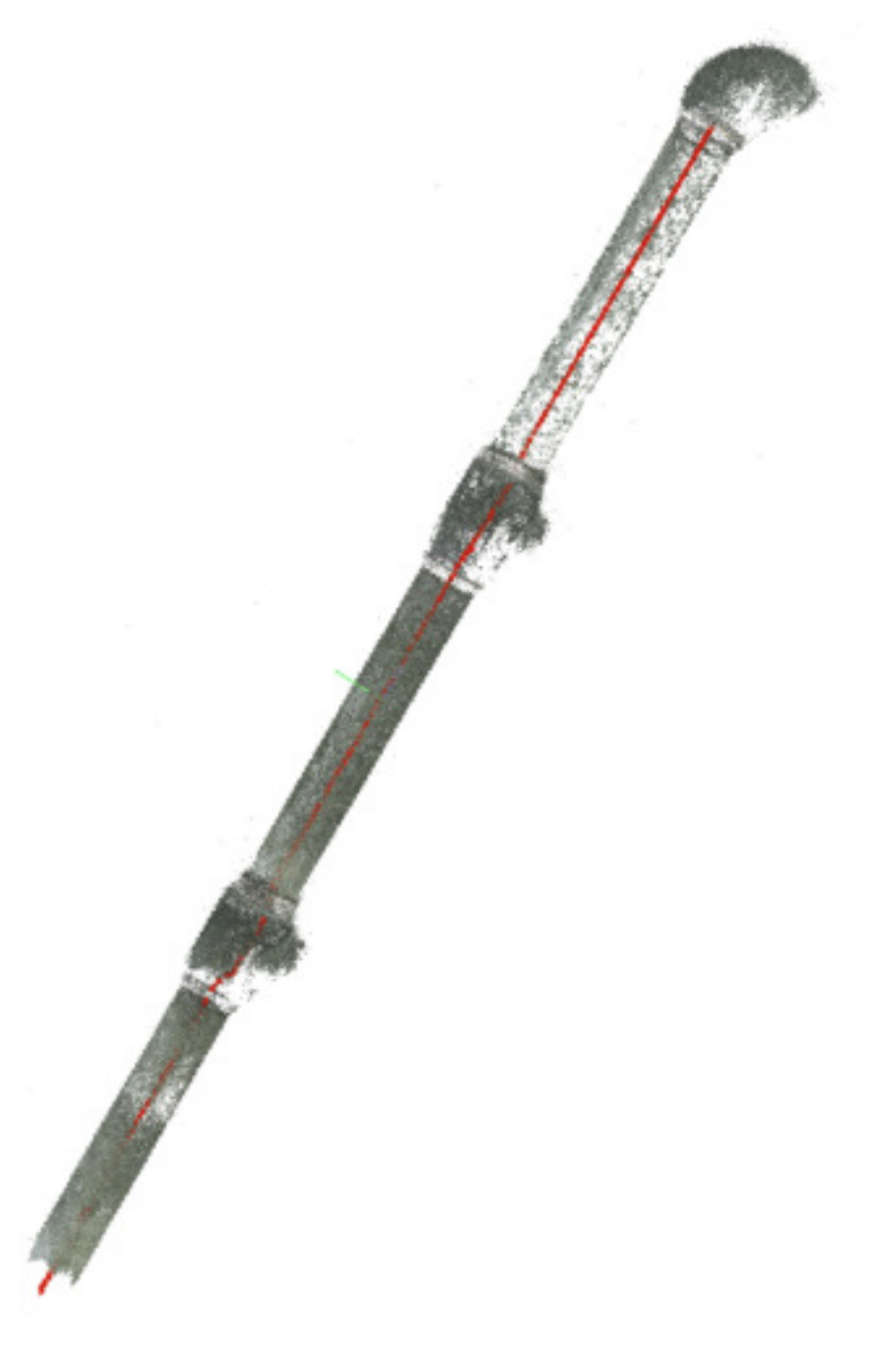}
         &
          \includegraphics[clip, height=\thisheight]
            {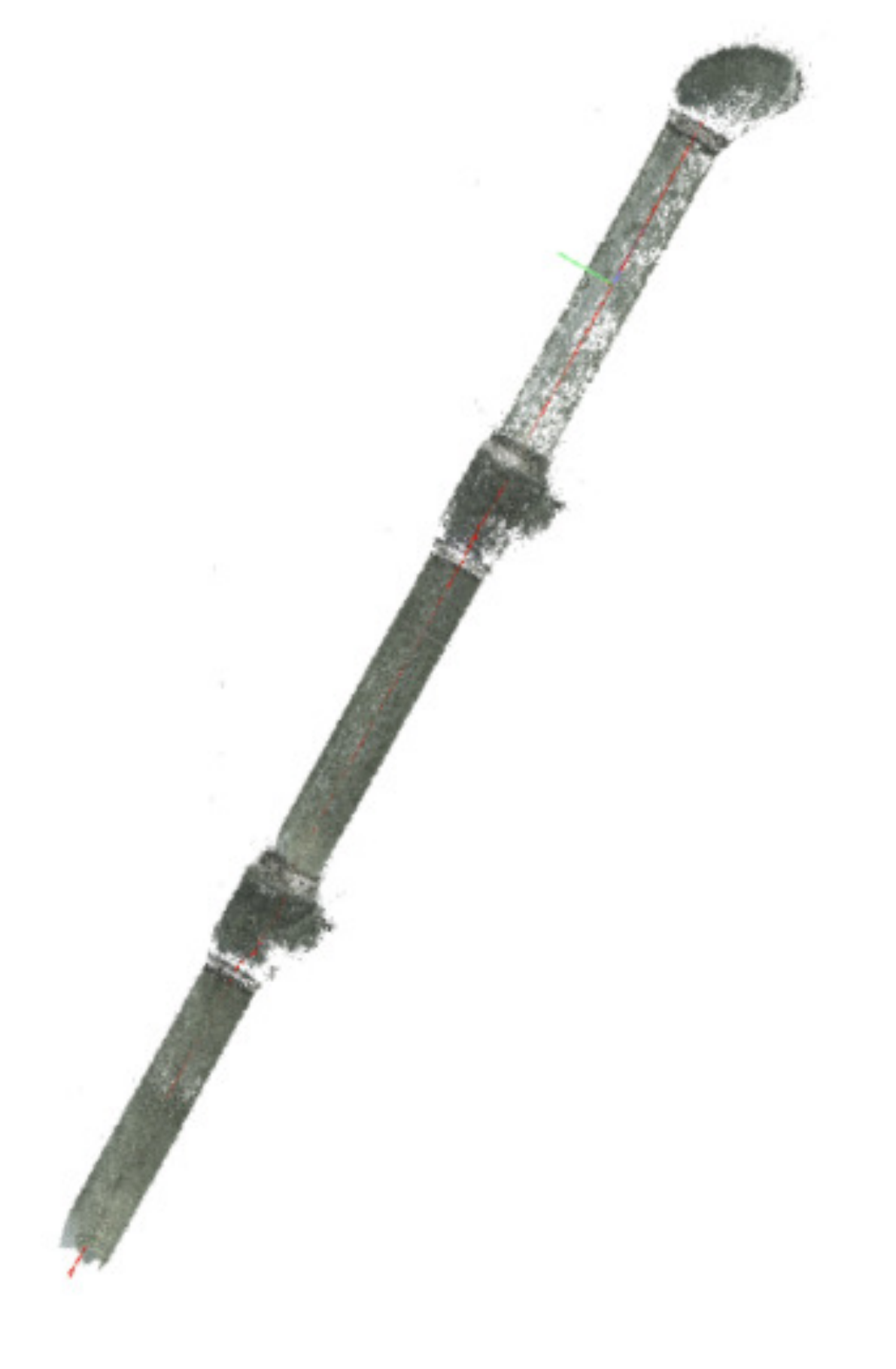}
         \\ 
        \rot{\rlap{~ \footnotesize \hspace{15pt} Network B}} &
        \includegraphics[height=\thisheight]{sequenceB_appearance_ds.png}
        &
            \includegraphics[clip, height=\thisheight]
            {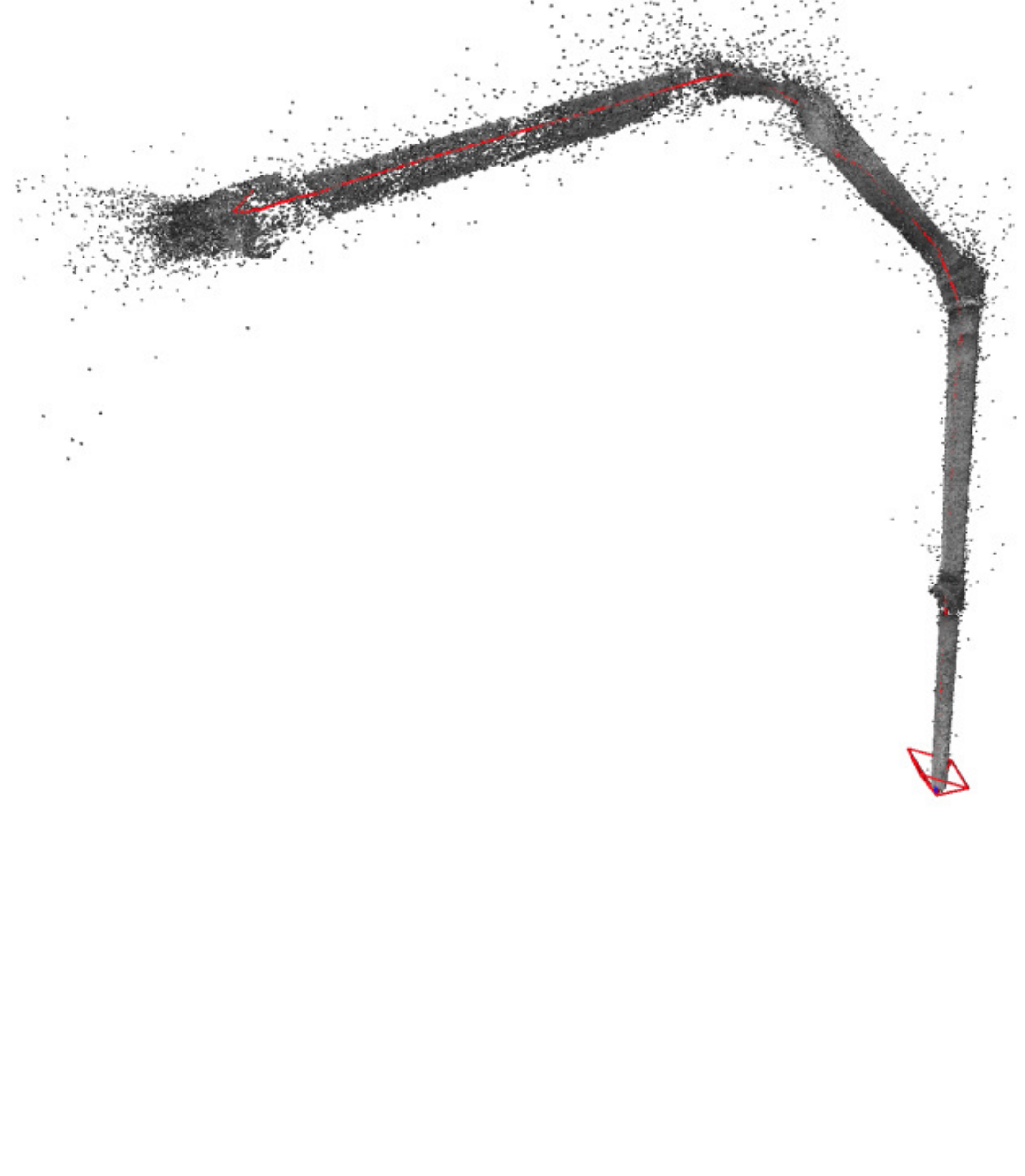}
         &
            \includegraphics[clip, height=\thisheight]
            {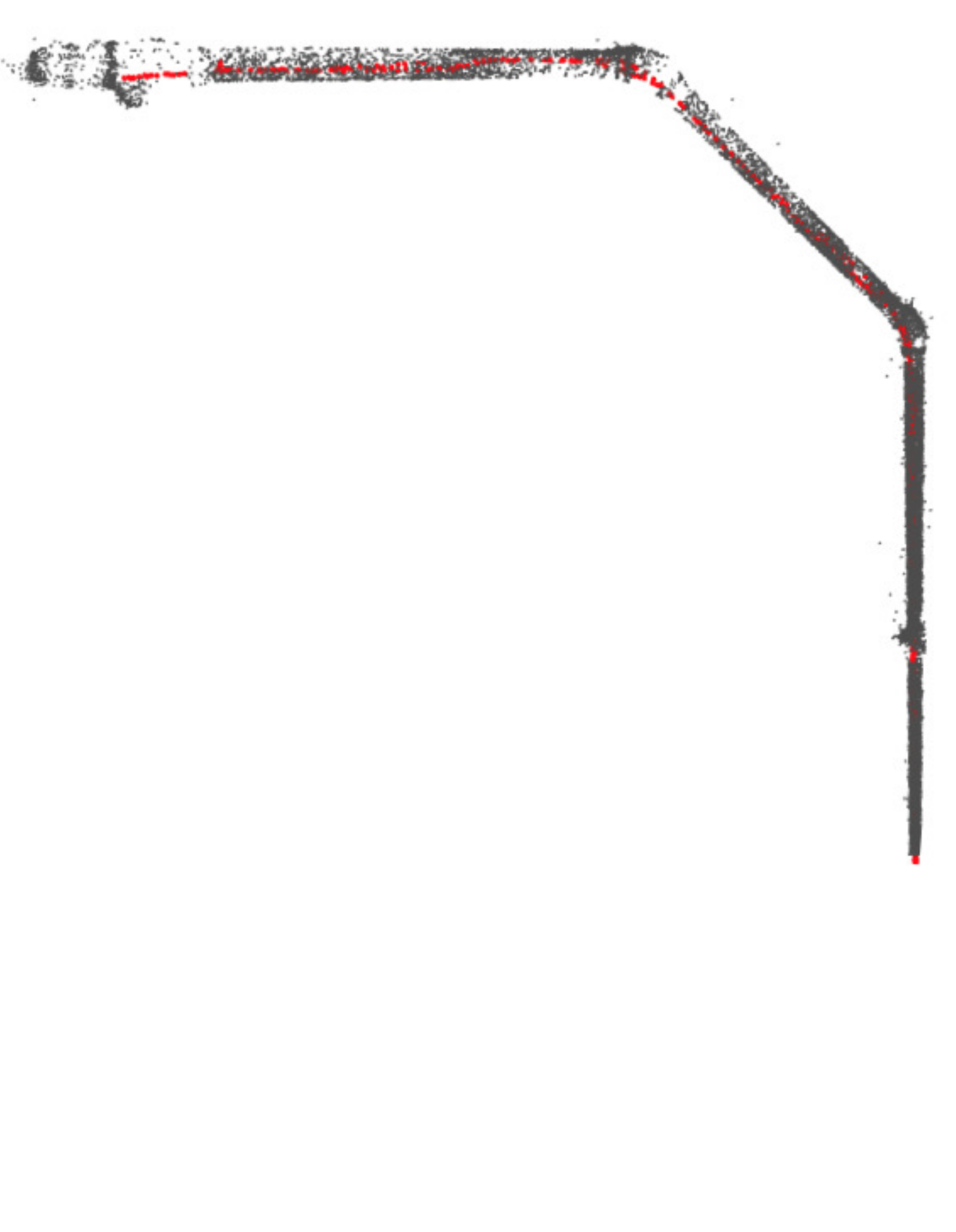}
         &
          \includegraphics[clip, height=\thisheight]
            {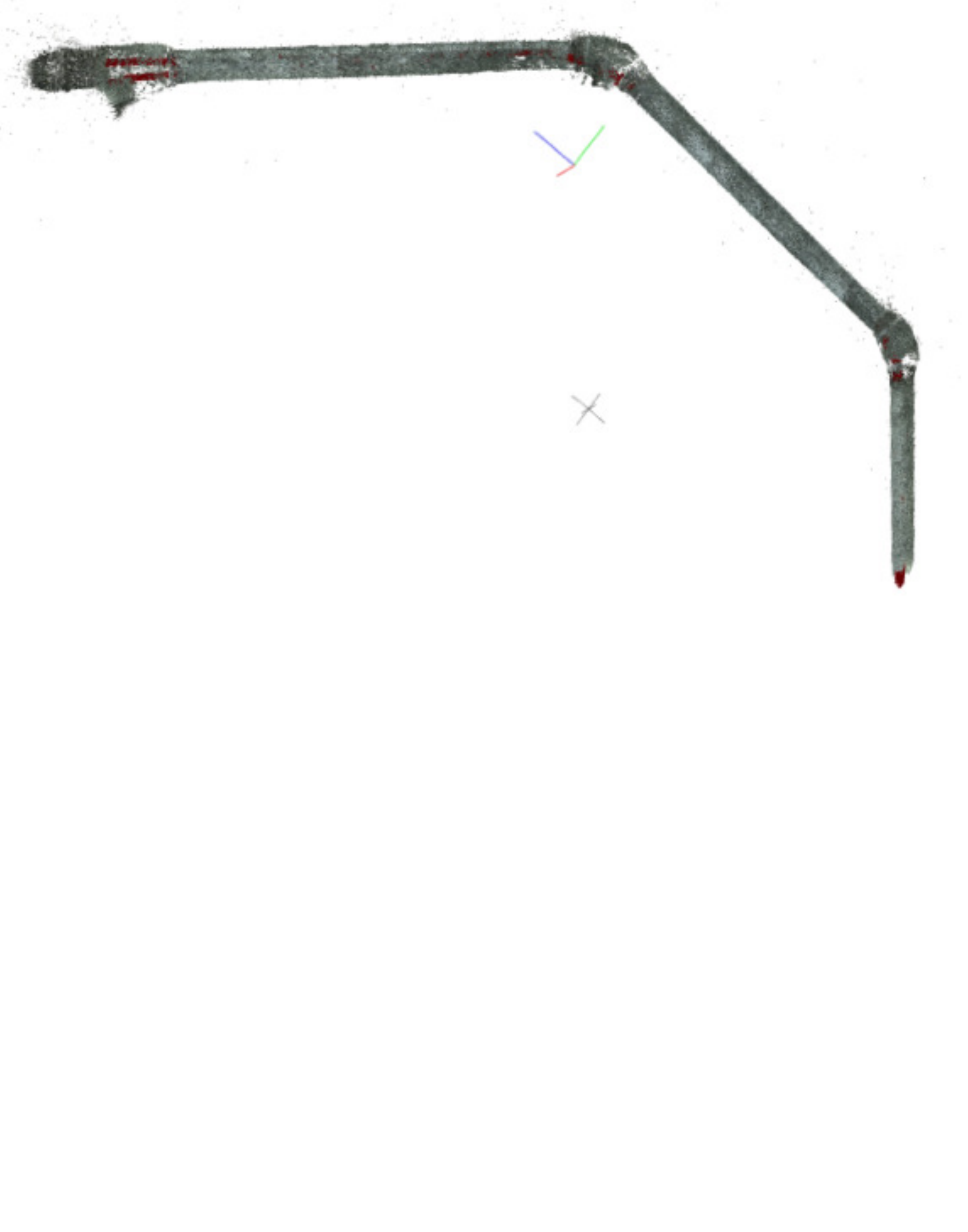}
         &
          \includegraphics[clip, height=\thisheight]
            {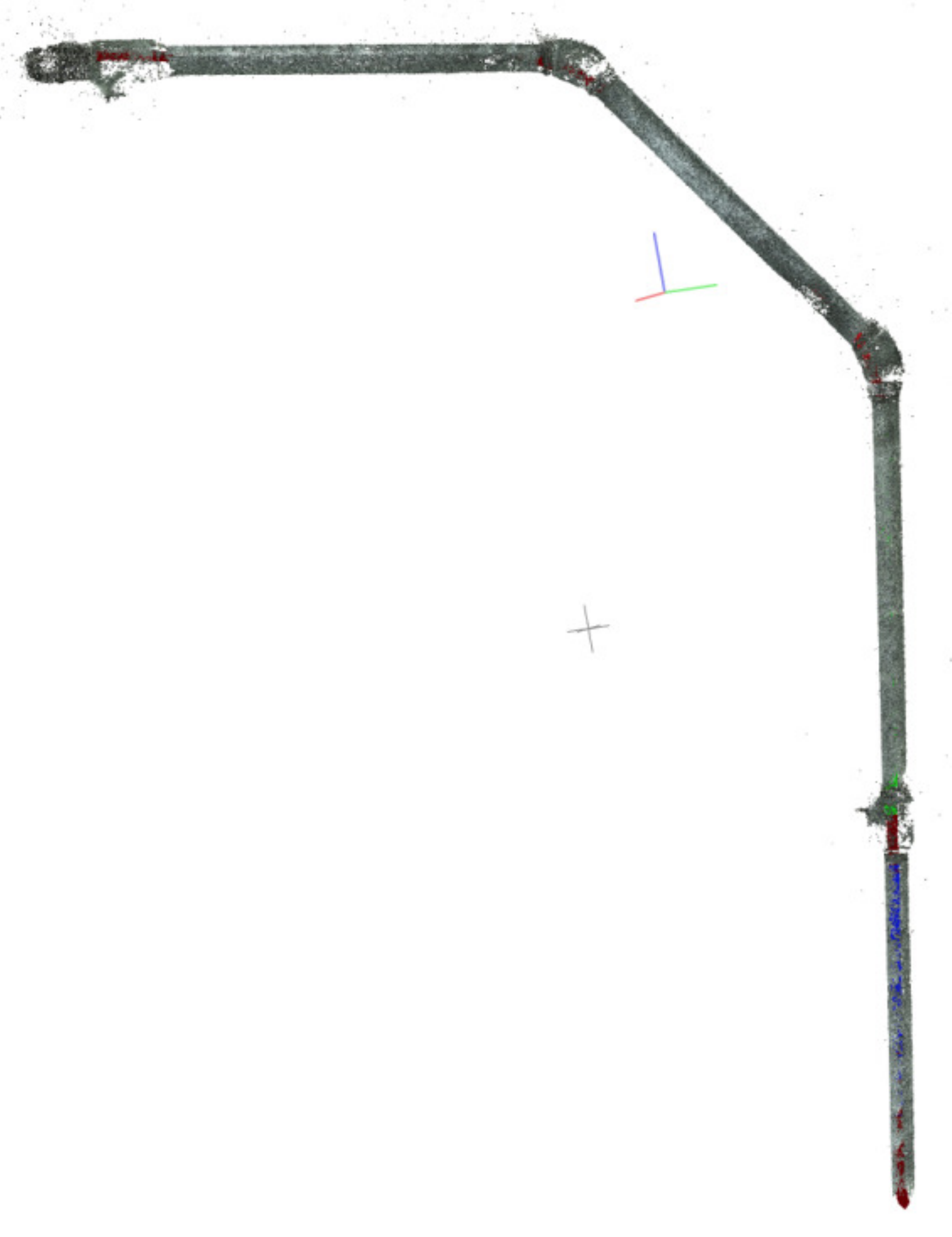}
         \\ 
        \rot{\rlap{~ \footnotesize \hspace{15pt} Network C}} &
        \includegraphics[height=\thisheight]{sequenceC_appearance_ds.png}
        &
            \includegraphics[clip, height=\thisheight]
            {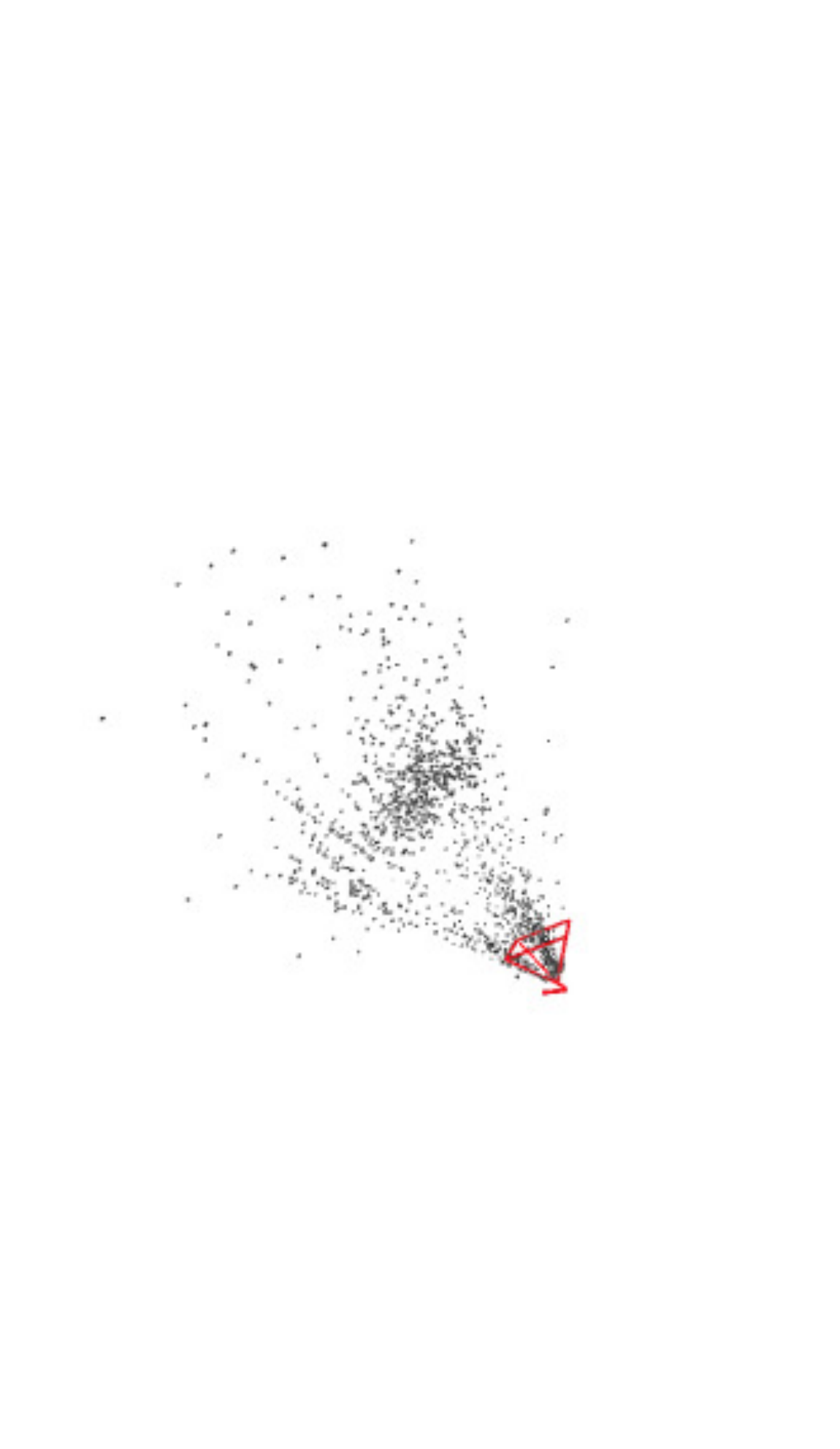}
         &
            \includegraphics[clip, height=\thisheight]
            {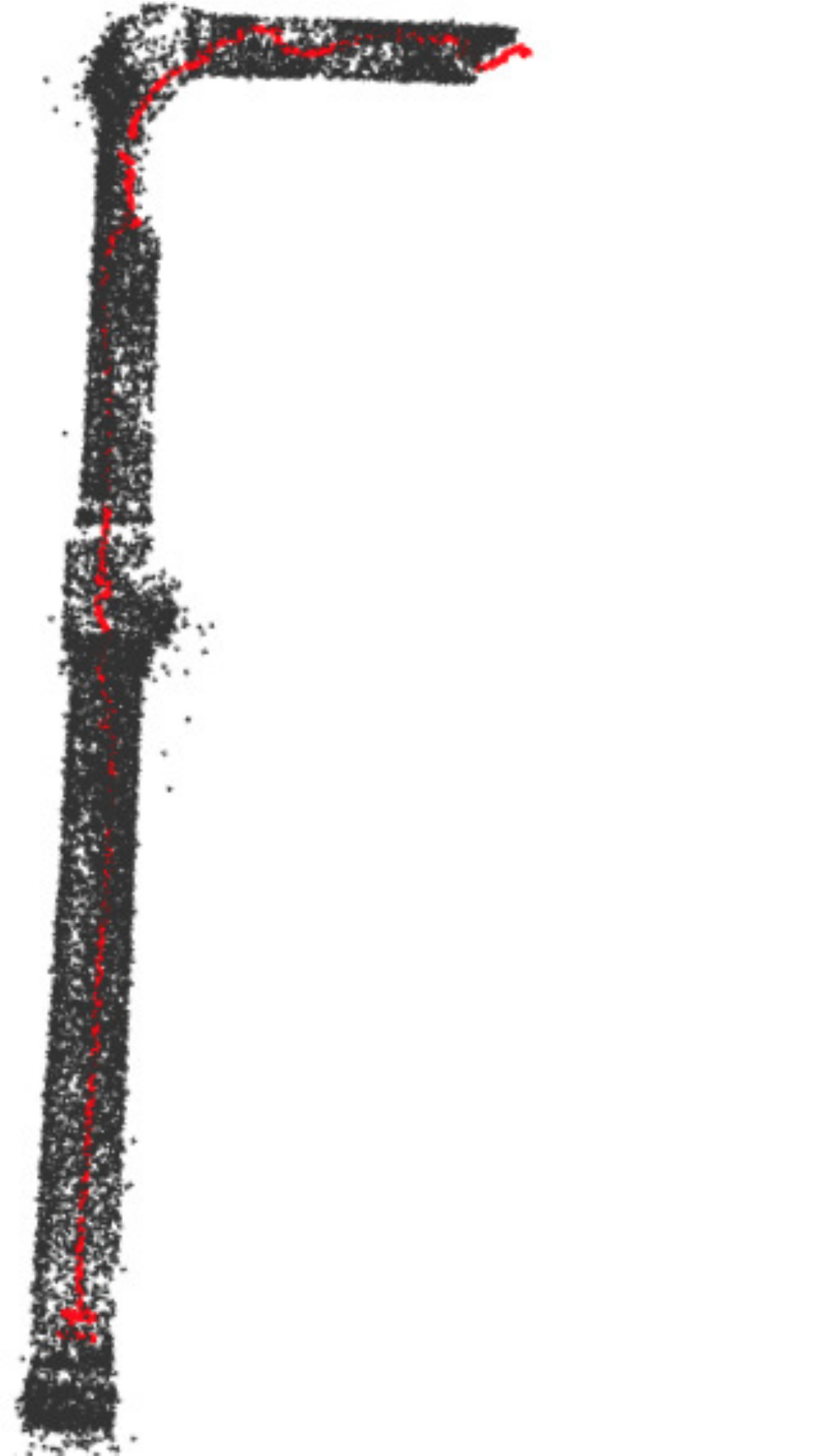}
         &
          \includegraphics[clip, height=\thisheight]
            {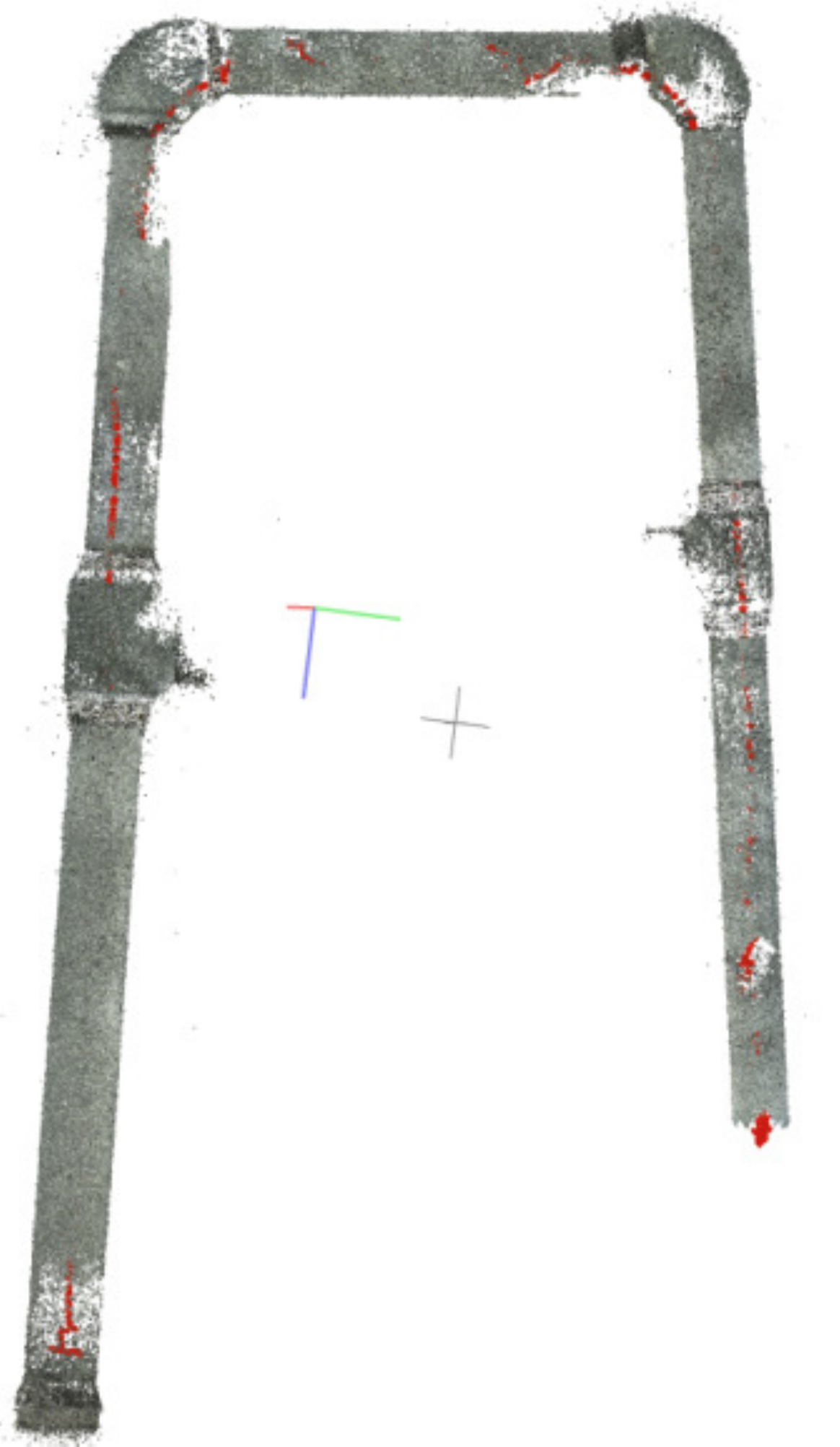}
         &
          \includegraphics[clip, height=\thisheight]
            {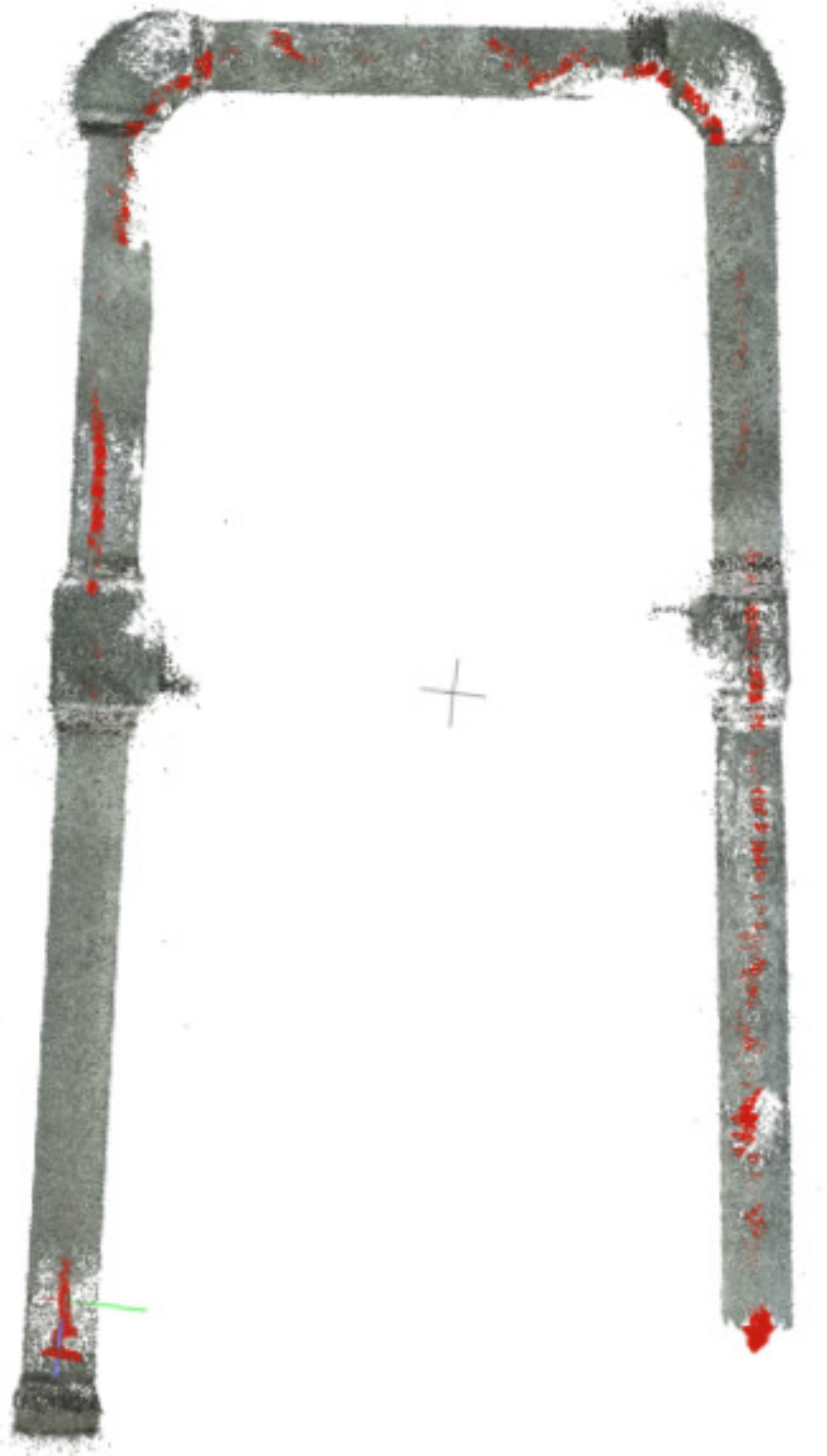}
        \\ 
        \rot{\rlap{~ \footnotesize \hspace{15pt} Network D}} &
        \includegraphics[height=\thisheight]{sequenceD_appearanceds.png}
        &
            \includegraphics[clip, height=\thisheight]
            {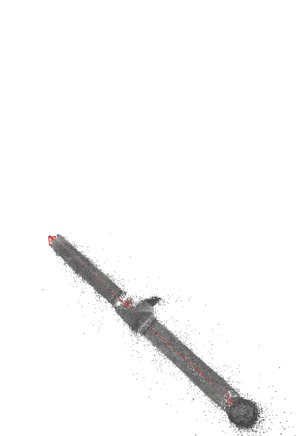}
         &
            \includegraphics[clip, height=\thisheight]
            {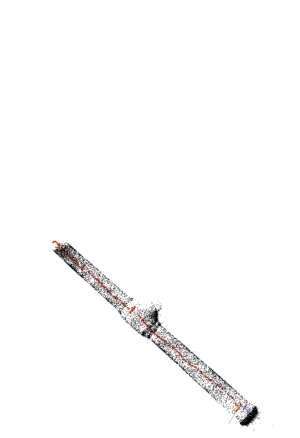}
         &
          \includegraphics[clip, height=\thisheight]
            {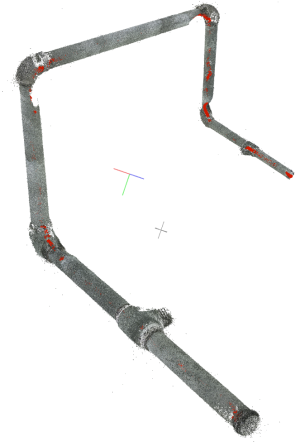}
         &
          \includegraphics[clip, height=\thisheight]
            {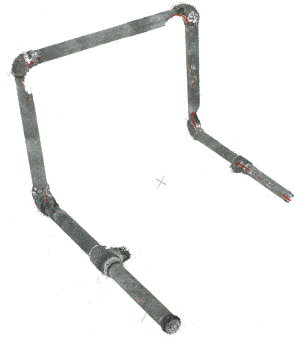}
        \\ \hline
        & 
        {\footnotesize Appearances}&
        {\footnotesize (a) DSO} &
        {\footnotesize (b) ORB-SLAM} &
        {\footnotesize (c) COLMAP} &
        {\footnotesize (d) {\bf Ours}} \\
      \end{tabular}
      \caption{{\bf Qualitative comparisons.} Each row shows a visual comparison of the 3D models for each pipe network obtained via four relevant methods. Gray dots show the reconstructed scene points, whereas red dots show the estimated cameras. }
      \label{fig:qualitative}
\end{figure*}
}

\section{Conclusion}
\noindent
In this paper, we have proposed a vision-based pipe reconstruction system that can provide an accurate 3D reconstruction for industrial endoscopic images. To deal with the accumulated model errors, our method incorporates the prior information of the pipe network without limiting the flexibility of camera motion. Proposed SfM pipeline consists of robust pipe detection and bundle adjustment constrained by the geometrical properties of a pipe system, which are carefully combined into an incremental image registration process, for stable camera tracking and 3D reconstruction. Throughout the experiments on realistic pipe network environments, it is demonstrated that our method can suppress scale drifting and reconstruct 3D pipe models more accurately and robustly than existing state-of-the-art methods. One of the future works is to develop a real-time application for giving instant feedback to the inspector. 

\para{Acknowledgement.} This work was partly supported by JSPS KAKENHI Grant Number 17H00744.

{\small
\bibliographystyle{IEEEtran}
\bibliography{shortstrings,references}
}

\end{document}